\documentclass[runningheads]{llncs}

\usepackage{eccv}

\usepackage{eccvabbrv}

\usepackage{graphicx}
\usepackage{booktabs}
\usepackage{makecell}

\usepackage[accsupp]{axessibility}  

\usepackage[dvipsnames]{xcolor}

\usepackage{algorithm}
\usepackage{algpseudocode}

\usepackage{placeins}

\algnewcommand\algorithmicinput{\textbf{Input:}}
\algnewcommand\Input{\item[\algorithmicinput]}
\algnewcommand\algorithmicoutput{\textbf{Output:}}
\algnewcommand\Output{\item[\algorithmicoutput]}
\algrenewcommand\algorithmicfunction{\textbf{Function:}}
\algrenewcommand\Function{\item[\algorithmicfunction]}
\algrenewcommand\algorithmicreturn{\textbf{Return:}}
\algrenewcommand\Return{\item[\algorithmicreturn]}

\usepackage{hyperref}

\usepackage{orcidlink}

\begin{document}

\title{Guide-and-Rescale: Self-Guidance Mechanism for Effective Tuning-Free Real Image Editing} 

\titlerunning{Guide-and-Rescale}

\author{Vadim Titov\inst{2*} \and Madina Khalmatova\inst{4*}  \and Alexandra Ivanova\inst{1,2,3*} \and Dmitry Vetrov\inst{5} \and Aibek Alanov\inst{1,2}}

\authorrunning{V.~Titov et al.}

\institute{HSE University \and
AIRI \\
\email{titov.vn@phystech.edu}, \email{\{a.ivanova,alanov\}@airi.net} \and
Skolkovo Institute of Science and Technology \and
UNSW Sydney \\
\email{m.khalmatova@student.unsw.edu.au} \and
Constructor University, Bremen\\
\email{dvetrov@constructor.university}}

\maketitle
\def\thefootnote{*}\footnotetext{Equal contribution.}
\begin{center}
    \centering
    \vspace{-0.5cm}
    \includegraphics[width=1\textwidth]{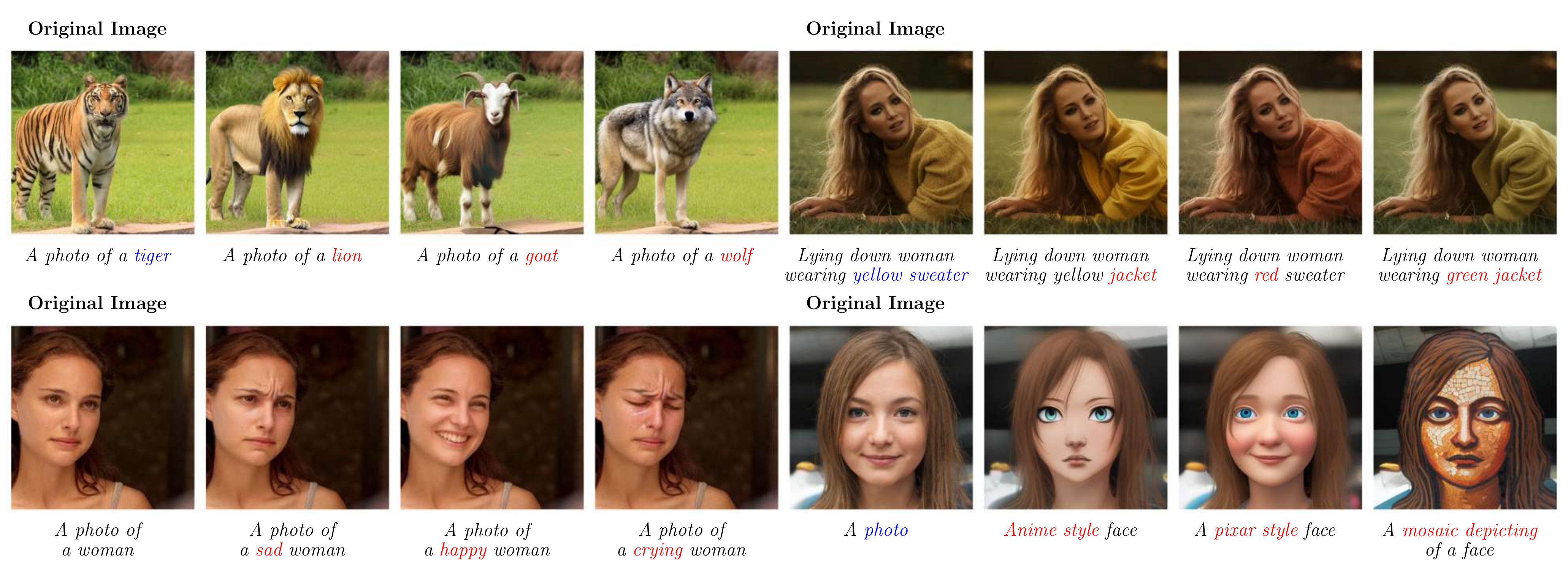} 
    \captionof{figure}{\textbf{Guide-and-Rescale for real image editing.} Our method allows to manipulate images for a wide range of different editings. It achieves a good balance between quality of manipulation and preservation of the original image.}
\label{fig:teaser}
\end{center}

\begin{abstract}

Despite recent advances in large-scale text-to-image generative models, manipulating real images with these models remains a challenging problem. The main limitations of existing editing methods are that they either fail to perform with consistent quality on a wide range of image edits or require time-consuming hyperparameter tuning or fine-tuning of the diffusion model to preserve the image-specific appearance of the input image. 
We propose a novel approach that is built upon a modified diffusion sampling process via the guidance mechanism. In this work, we explore the self-guidance technique to preserve the overall structure of the input image and its local regions appearance that should not be edited. In particular, we explicitly introduce layout-preserving energy functions that are aimed to save local and global structures of the source image. Additionally, we propose a noise rescaling mechanism that allows to preserve noise distribution by balancing the norms of classifier-free guidance and our proposed guiders during generation. 
Such a guiding approach does not require fine-tuning the diffusion model and exact inversion process. As a result, the proposed method provides a fast and high-quality editing mechanism.
In our experiments, we show through human evaluation and quantitative analysis that the proposed method allows to produce desired editing which is more preferable by humans and also achieves a better trade-off between editing quality and preservation of the original image.
Our code is available at \href{https://github.com/MACderRu/Guide-and-Rescale}{https://github.com/MACderRu/Guide-and-Rescale}.

\end{abstract}    

\section{Introduction}
\label{sec:intro}

In recent years, diffusion models \cite{NEURIPS2021_5dca4c6b, ho2020denoising} have been rapidly developed due to their high generation quality.
Therefore, they have started to be actively used as a base model for text-to-image generative models \cite{ramesh2022hierarchical, rombach2022high, saharia2022photorealistic}, where an image has to be generated from a textual description. Although such models have already achieved impressive results,
they are still difficult to apply to the task of editing real images. The main problem is to find a balance between making the edited image reflect the expected change and preserving the original structure and the parts that should not have been edited.

Existing approaches to image manipulation based on text-to-image models address this task in different ways and can be categorized into three main groups of methods. Approaches from the first category \cite{kawar2023imagic, zhang2022sine, Valevski2022UniTuneTI} use fine-tuning of the whole diffusion model on the input image in order to preserve its structure and all necessary details during editing. Although such a strategy gives good results, the optimization process makes them very long, which does not allow to apply them in practice. 
In the next group of methods \cite{hertz2022prompt, 10.1145/3588432.3591513, Tumanyan_2023_CVPR, brooks2022instructpix2pix, cao_2023_masactrl, han2023improving, epstein2023selfguidance}, instead of the optimization step, they use internal representations of images in the diffusion model and replace them during generation to preserve parts of the original image that should not change during editing. These methods are significantly faster than optimization-based methods, but they are not universal and require careful tuning of hyperparameters. As a result, they can only work consistently for a narrow set of edits. Finally, the third group of methods \cite{mokady2022null, miyake2023negativeprompt, 10204740, pan2023effective} focuses on building a high-quality reconstruction of the original image by minimizing the discrepancy between the forward and backward trajectories of the diffusion processes. The main problem of this group of methods is the additional time required to construct a good quality inversion. 

In general, the currently available methods for manipulating real images have many limitations in terms of time, quality, and edit versatility, so it is still a challenging problem to find a model that is efficient, high quality, and covers most of the edit types.

In this paper, we investigate the guidance technique for the problem of real image editing. As most of the tunning-free approaches leverage UNet internal representations which are cached during inversion and used during editing generation, it causes misalignment between real features and inserted ones, and therefore noise trajectory may leave the distribution of natural images. To overcome these issues we propose special energy functions named \textit{guiders} that are designed to preserve overall source image structure and local regions that should not be edited as well. We explicitly introduce preserving term that smoothly manipulates overall noise trajectory distribution. Additionally, we extend our framework by the automatic mechanism of noise rescaling which prevents discrepancy of noise trajectory from initial diffusion sampling by balancing the norms of classifier-free guidance and our proposed guiders. Our approach does not require the exact reconstruction as well as the fine-tuning the diffusion model, so it is computationally efficient. Due to the flexibility of the proposed guiders, our method is versatile and supports many types of editing, from local changes of objects to global stylisation (see Fig. \ref{fig:teaser}).

We apply our method to Stable Diffusion \cite{rombach2022high} model and conduct an extensive set of experiments for comparison with other methods. Our approach achieves a better balance between editing quality and preservation of the original image structure in terms of CLIP/LPIPS scores for a wide range of different editing types. For a standard image-to-image problem (Dog $\rightarrow$ Cat) we show that our method performs better than other baselines. Also, we provide user studies and show that our approach is most preferable in human evaluation.

\section{Related Work}
\label{sec:related_work}

Diffusion models \cite{NEURIPS2021_5dca4c6b, ho2020denoising} show high-quality results in generative modeling. This is the reason why they could be used to solve the image editing problem. Text-to-image diffusion models \cite{ramesh2022hierarchical, rombach2022high, saharia2022photorealistic} are particularly useful and common for solving this problem. One of the first text-guided image editing methods is SDEdit \cite{meng2022sdedit} whose main idea is to add the noise step by step and then denoise with different conditions, but the quality of the editing is not so high. We can distinguish three main groups of editing methods which we consider next.

\textbf{Optimization-based methods.}
Methods in the first group \cite{kawar2023imagic, zhang2022sine, Valevski2022UniTuneTI} are based on fine-tuning the diffusion for the input image. In Imagic \cite{kawar2023imagic} and SINE \cite{zhang2022sine} the diffusion model is fine-tuned on the original image, and the target prompt, which allows taking into account initial image characteristics but slows down the editing process. UniTune \cite{Valevski2022UniTuneTI} fine-tunes the model only on the original image, but it still works slowly. However, UniTune, in contrast to Imagic and SINE, allows applying multiple edits to one image without any additional optimization.

\begin{figure*}[t]
    \centerline{\includegraphics[width=\textwidth]{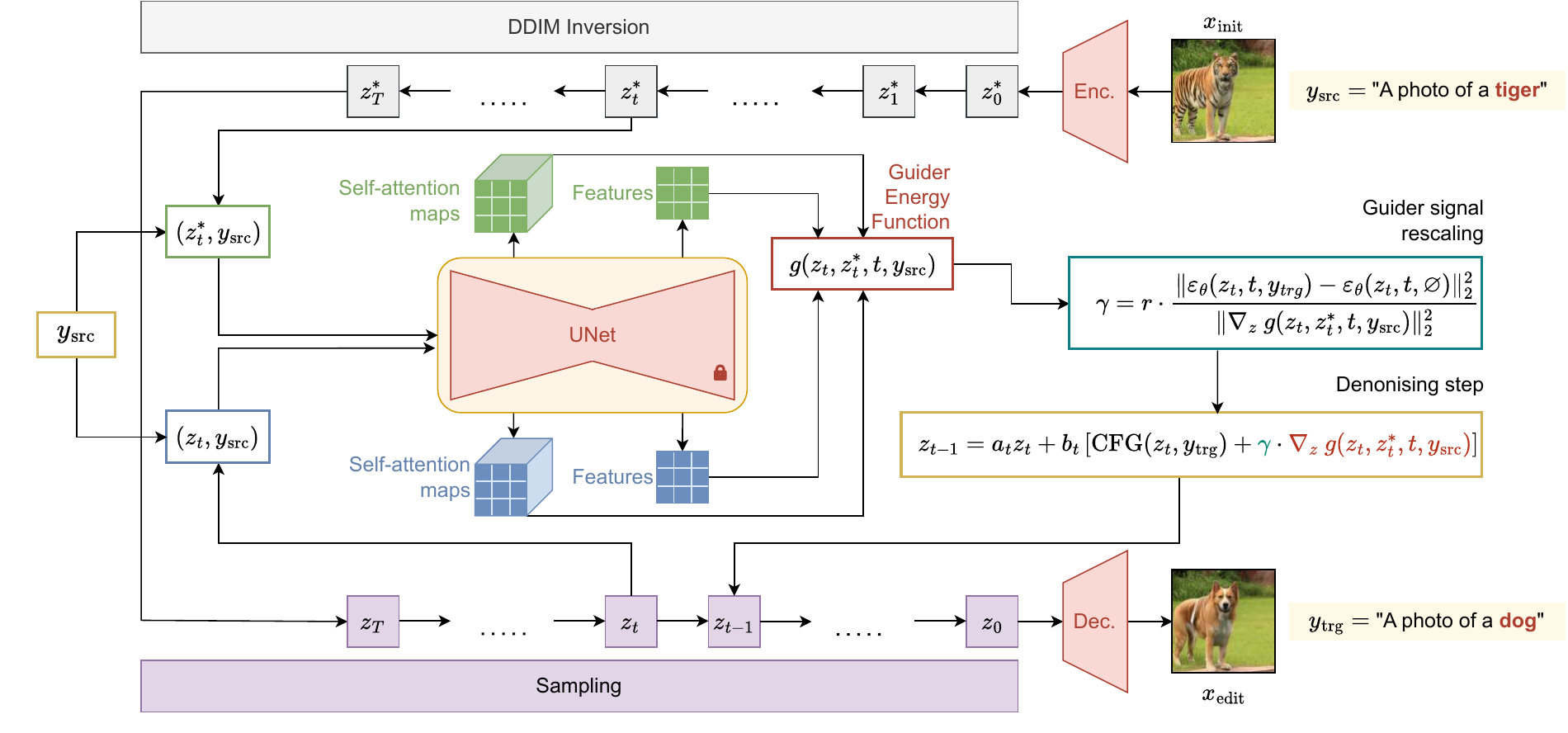}}
	\caption{Overall scheme of the proposed method \textit{Guide-and-Rescale}. First, our method uses a classic ddim inversion of the source real image. Then the method performs real image editing via the classical denoising process. For every denoising step the noise term is modified by guider that utilizes latents $z_t$ from the current generation process and time-aligned ddim latents $z^*_t$.}
	\label{fig:method_diagram}
\end{figure*}

\textbf{Methods that utilize internal representations of the diffusion model.}
The following approaches \cite{hertz2022prompt, 10.1145/3588432.3591513, Tumanyan_2023_CVPR, brooks2022instructpix2pix, cao_2023_masactrl, han2023improving, epstein2023selfguidance} are based on using internal representations of the image in the text-to-image diffusion model, such as features from cross-attention and self-attention layers to preserve the structure and the details of the original image.
Prompt-to-Prompt \cite{hertz2022prompt} controls the cross-attention by replacing maps for new tokens from the target prompt and preserving maps in the overlapping tokens. In \cite{10.1145/3588432.3591513} authors guide cross-attention maps to reference reconstruction with a source prompt during the denoising process. Plug-and-Play \cite{Tumanyan_2023_CVPR} outperforms Prompt-to-Prompt by using feature and self-attention outputs as conditions for the diffusion model to focus on spatial features and their self-affinities. InstructPix2Pix \cite{brooks2022instructpix2pix} proposes an approach based on a combination of Prompt-to-Prompt and the language model to control editing through specific instructions, but it requires an additional dataset generation phase. 

The limitation of previous methods is the difficulty of changing poses and object locations. MasaCtrl \cite{cao_2023_masactrl} uses mutual self-attention features based on reconstruction during editing to preserve the local content and textures of the original image. In ProxMasaCtrl \cite{han2023improving}, which is MasaCtrl extension, an additional proximal gradient step for the scaled noise difference is used to change the geometry and layout in edited images. Despite the ability to change object poses, the disadvantage of MasaCtrl is that the original image, the object background, is preserved inaccurately. Self-guidance \cite{epstein2023selfguidance} proposes an approach to simple editings such as changing object position, size, and color by using intermediate activations and attention interactions as guidance signals during generation. The method we propose is based on this self-guidance mechanism, but we develop it for any type of editing.

\textbf{Methods that improve the inversion part.}
The next group of methods \cite{mokady2022null, miyake2023negativeprompt, 10204740, pan2023effective} focuses on improving the quality of the image reconstruction, and it allows to preserve parts of the image that should not be edited and the overall structure. This is achieved by narrowing the gap between the forward and backward trajectories of the diffusion model. 
Null-text Inversion \cite{mokady2022null} is proposed as an adaptation of Prompt-to-Prompt \cite{hertz2022prompt} for real images and is based on the optimization of the null-text embedding which is used in classifier-free guidance. To avoid learning the null-text embedding for each image, a new method for its approximation by the source prompt embedding is suggested in Negative-prompt Inversion \cite{miyake2023negativeprompt}, but it can lead to artifacts. EDICT \cite{10204740} overcomes the restrictions of DDIM \cite{song2021denoising} in the image reconstruction by mathematically precise image inversion, but requires some additional computation time. Finally, AIDI \cite{pan2023effective} is proposed to increase the precision of the reconstruction by applying fixed-point iteration, Anderson acceleration in inversion, and blended guidance for sampling. 

The main problem of the methods in this group is the extra time needed to achieve a high-quality reconstruction of the initial image. Our method in contrast does not require an optimization and additional phase for image reconstruction, and therefore it is more computationally efficient.

\section{Method}
\label{sec:method}

\subsection{Preliminaries}\label{sec:method_preliminaries}

\noindent\textbf{Diffusion Model.} The method uses a pre-trained text-to-image diffusion model. In our experiments, we use a publicly available Stable Diffusion (SD) \cite{rombach2022high} model, specifically its checkpoint \textit{stable-diffusion-v1-4}. Our method does not finetune or modify it.

SD is a latent diffusion model (LDM) \cite{rombach2022high}, meaning that the model operates in a latent space. Using a pre-trained VAE \cite{Kingma2014} encoder, $Enc.$, one can encode an image into a sample from the latent space. And, likewise, given a latent $z$ and a VAE decoder, $Dec.$, one can obtain an image space sample $x$. However, our method does not rely on the existence of the latent space, so it may be applied to diffusion models, that perform sampling directly in the image space \cite{saharia2022photorealistic}.

SD is a text-to-image model, meaning that it can be conditioned on textual captions. Our method heavily relies on this property, as it requires two prompts to be specified: a source prompt, describing the initial image, and a target prompt, outlining the desired result of editing.

We use DDIM \cite{song2021denoising} sampling with $T = 50$ intermediate steps, where a single sampling step is defined as:

\vspace{-0.4cm}
\begin{equation}\label{eq:ddim_sampling}
    \begin{array}{c}
        z_{t-1} =  a_tz_t + b_t \varepsilon_{\theta}(z_t, t, y),
    \end{array}
\end{equation}
\noindent where $a_t, b_t$ are specified coefficients, $z_t$ is a current latent, $t$ is a current timestep, $y$ is conditioning data and $\varepsilon_{\theta}(z_t, t, y)$ is the noise, predicted by the diffusion model.

To allow the diffusion model to operate with real images, we need a noised latent $z_T$, corresponding to this image. For this purpose, we use DDIM inversion, defined similarly to Equation \ref{eq:ddim_sampling}:

\vspace{-0.4cm}
\begin{equation}\label{eq:ddim_inversion}
    \begin{array}{c}
        z_{t+1} =  a^*_tz_t + b^*_t \varepsilon_{\theta}(z_t, t, y).
    \end{array}
\end{equation}

Even though DDIM inversion is not stable, as noted by the authors of Prompt-to-Prompt \cite{hertz2022prompt}, our method overcomes this issue without any enhancements to the inversion process or optimization and is capable of preserving features of the initial image, as we show in experiments.

\noindent\textbf{Guidance.} Guidance mechanism takes the prediction of the diffusion model and enhances it with information from additional data. Classifier guidance relies on a separate model, classifier $p(y|z_t)$, trained on noised latents. As long as this classifier is differentiable with respect to the current latent, we can subtract score function $\nabla_{z_t} \log p(y|z_t)$ from the diffusion model prediction to sample from the conditional distribution $p(z_t|y)$ instead of the data distribution $p(z_t)$.

Training a separate classifier on specific data may sometimes be a resourceful task. Given a conditional diffusion model, one can rely on this model’s knowledge about additional data, that it inquires with conditioning capabilities. This guidance technique is called classifier-free guidance (CFG) \cite{ho2022classifierfree}. To determine the sampling direction, which leads to correspondence with conditioning data $y$, CFG compares a conditional prediction of the model with an unconditional one. In case of text-to-image models, the latter can be easily obtained by conditioning the model on the empty text $\varnothing = ``"$. With CFG incorporated, the diffusion model prediction, used in both sampling, Equation \ref{eq:ddim_sampling}, and inversion, Equation \ref{eq:ddim_inversion}, takes the form of $\hat{\varepsilon}_{\theta}(z_t, t, y)$:

\vspace{-0.4cm}
\begin{equation}\label{eq:cfg}
    \begin{array}{c}
        \hat{\varepsilon}_{\theta}(z_t, t, y) = \mathrm{CFG}(z_t, t, y, w)
        = \varepsilon_{\theta}(z_t, t, \varnothing) + w \big( \varepsilon_{\theta}(z_t, t, y) \! - \! \varepsilon_{\theta}(z_t, t, \varnothing) \big),
    \end{array}
\end{equation}

\noindent where $w$ is a guidance scale, that controls to which extent additional data $y$ influences the generation process. For SD model, the guidance scale is typically chosen as $w = 7.5$. It is important to note, that with $w=1$ CFG sampling step equals regular conditional sampling.

\noindent\textbf{Self-guidance.} As proposed in \cite{epstein2023selfguidance}, the choice in guidance sources is not limited to either the classifier or the diffusion model itself. One can use any energy function $g$ to guide the sampling process, as long as there exists a gradient with respect to $z_t$. When the energy function uses outputs of internal layers of the diffusion model, the guidance process is called \textit{self-guidance}. The authors of the method suggest defining energy function $g$ on top of cross-attention maps $\mathcal{A}^{\mathrm{cross}} := \mathrm{cross\;attn.}[\varepsilon_{\theta}(z_t, t, y)]$ and the output of penultimate layer of the diffusion model decoder, features $\Psi := \mathrm{features}[\varepsilon_{\theta}(z_t, t, y)] $. To add self-guidance to CFG, one has to modify $\hat{\varepsilon}_{\theta}(z_t, t, y)$ from Equation \ref{eq:cfg} as:

\vspace{-0.3cm}
\begin{equation}\label{eq:sampling_base}
    \begin{array}{c}
        \hat{\varepsilon}_{\theta}(z_t, t, y) = \mathrm{CFG}(z_t, t, y, w) + v\cdot \nabla_{z_t} g(z_t, t, y, \mathcal{A}^{\mathrm{cross}}, \Psi),
    \end{array}
\end{equation}
\noindent where $v$ is a self-guidance scale.

\subsection{Guide and Rescale}\label{sec:method_guide_and_resacle}

\begin{figure}[t]
    \centerline{\includegraphics[width=\linewidth]{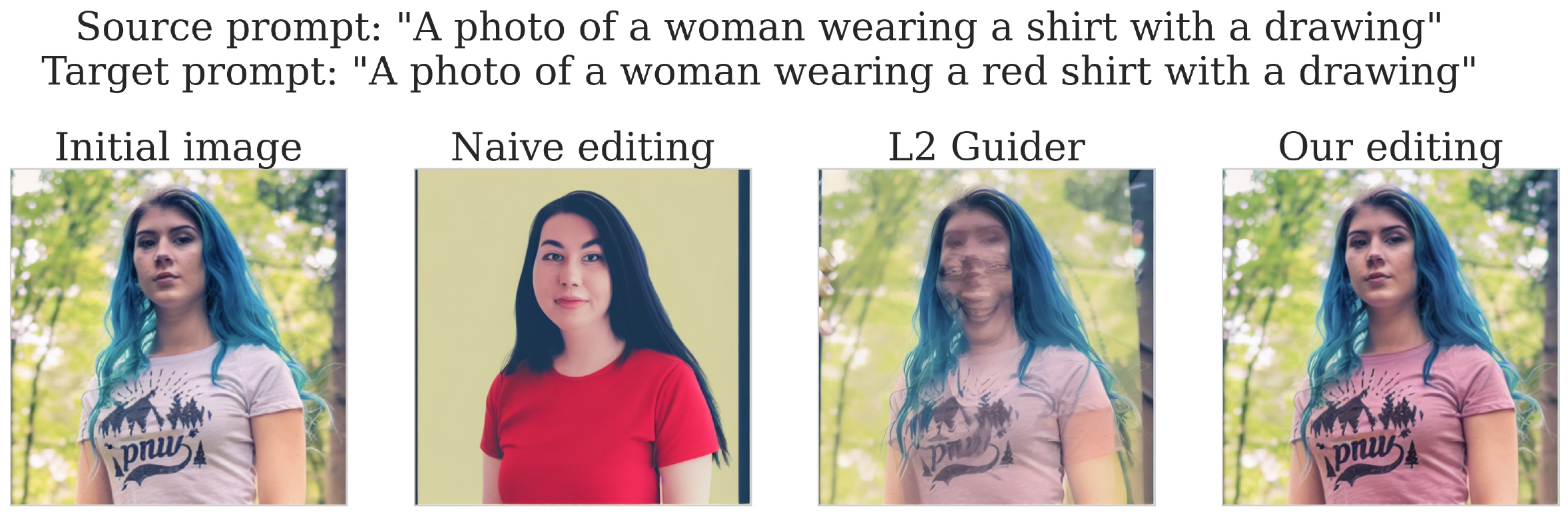}}
	\caption{Editing example. From left to right, first: initial image; second: naive editing, described in Equation \ref{eq:sampling_naive}; third: editing with simple energy function $g$ from Equation \ref{eq:simple_g}; fourth: editing with the proposed method.}
	\label{fig:example}
\end{figure}

Consider an initial image $x_{\mathrm{init}}$, a source prompt $y_{\mathrm{src}}$ and a target prompt $y_{\mathrm{trg}}$. For example, let $x_{\mathrm{init}}$ be a photo of a woman with blue hair, wearing a shirt with a drawing. A source prompt, describing the initial image, can be defined as $y_{\mathrm{src}} =``$A photo of a woman wearing a shirt with a drawing$"$. A target prompt should describe the desired editing result. For instance, if we want to change the color of the shirt to red, the target prompt can be defined as $y_{\mathrm{trg}} = ``$A photo of a woman wearing a \textit{red} shirt with a drawing$"$. This example is illustrated in Fig. \ref{fig:example}.

Our goal is to generate an image $x_{\mathrm{edit}}$, that would correspond to the editing, specified in $y_{\mathrm{trg}}$. At the same time, $x_{\mathrm{edit}}$ should contain all the features of $x_{\mathrm{init}}$, not influenced by editing. For example, when changing the color of the shirt, $x_{\mathrm{edit}}$ and $x_{\mathrm{init}}$ should be the same in all other aspects. Meaning, that we should preserve the background of the photo and the visual appearance of the woman, including the color of her hair, her facial expression, and her position.

A naive approach to obtain correspondence to $y_{\mathrm{trg}}$ can be described as follows. First, we obtain a DDIM inversion trajectory $\{z^*_t\}_{t=0}^T$ for $x_{\mathrm{init}}$, conditioning on $y_{\mathrm{src}}$. We use CFG with guidance scale $w=1$, i.e. regular conditional sampling:

\vspace{-0.3cm}
\begin{equation}\label{eq:inversion_process}
\begin{array}{c}
     z^*_0 = Enc.(x_{\mathrm{init}})\\
     z^*_{t+1} = a^*_t z^*_t + b^*_t \varepsilon_{\theta}(z^*_t, t, y_{\mathrm{src}}) \big|_{0 \le t \le T-1}.
\end{array}
\end{equation}

A noised latent $z^*_T$ has information about $x_{\mathrm{init}}$ encoded, so an obvious further step would be to synthesize an image, starting from obtained $z^*_T$ and conditioning on $y_{\mathrm{trg}}$, using CFG with standard sampling guidance scale $w=7.5$:

\vspace{-0.3cm}
\begin{equation}\label{eq:sampling_naive}
\begin{array}{c}
     \hat{z}_T = z^*_T\\
     \hat{z}_{t-1} = a_t \hat{z}_{t} + b_t \mathrm{CFG}(\hat{z}_{t}, t, y_{\mathrm{trg}}, 7.5)\big|_{1 \le t \le T}\\
     x_{\mathrm{edit}} = Dec.(\hat{z}_{0}).\\
\end{array}
\end{equation}

This method achieves near-perfect coherence with $y_{\mathrm{trg}}$. However, due to a mismatch of guidance scales and conditioning data in inversion and sampling processes, trajectory $\{\hat{z}_t\}_{t=0}^T$ ends up being too far from  $\{z^*_t\}_{t=0}^T$. This results in $x_{\mathrm{init}}$ being modified significantly, as shown by the second image in Fig. \ref{fig:example}.

In previous works, there are different workarounds suggested. For example, substituting inner representations of the diffusion model with corresponding representations from $\{z^*_t\}_{t=0}^T$ \cite{cao_2023_masactrl, Tumanyan_2023_CVPR, hertz2022prompt}, finetuning the model \cite{kawar2023imagic} or optimizing null-text embeddings \cite{mokady2022null} to move $\{\hat{z}_t\}_{t=0}^T$ closer to the inversion trajectory.

In this work, we suggest addressing this problem with modified self-guidance. As inversion trajectory is an almost perfect reconstruction trajectory for $x_{\mathrm{init}}$, we suggest guiding the sampling process with respect to $\{z^*_t\}_{t=0}^T$. This way we will be able to preserve features of the initial image. For example, we can simply define the function $g$ as an L2 norm of the difference of the current latent, obtained with CFG sampling in Equation \ref{eq:sampling_naive}, and the corresponding inversion latent:

\vspace{-0.3cm}
\begin{equation}\label{eq:simple_g}
    g(\hat{z}_t, z^*_t) = \|\hat{z}_t - z^*_t\|^2_2.
\end{equation}

However, for proper editing, it is important to control, which regions of the image we want to preserve with the function $g$, and which areas we still want to vividly edit. Based on editing results (Fig. \ref{fig:example}), the simple function in Equation \ref{eq:simple_g} does not hold this property.

We found that applying guidance over inner representations of the model offers the desired control over image regions. To obtain this inner representation, we need to conduct two separate forward passes over the diffusion model. 

First, for the inversion trajectory, we make a single sampling step with the same parameters, as in Equation \ref{eq:inversion_process}:

\vspace{-0.3cm}
\begin{equation}\label{eq:sampling_rec}
\begin{array}{c}
     \bar{z}^*_{t-1} = a_t z^*_t + b_t \varepsilon_{\theta}(z^*_t, t, y_{\mathrm{src}}).
\end{array}
\end{equation}
\noindent Together with a new latent $\bar{z}^*_{t-1}$, we obtain inner representations of the diffusion model $\varepsilon_{\theta}$, corresponding to the inversion trajectory. For now, we denote them as $\mathcal{I}^*$, i.e. $\mathcal{I}^*$ is a set of inner representations (for example, outputs of cross-attention or self-attention layers, etc.) that are calculated during the forward pass of $\varepsilon_{\theta}(z^*_t, t, y_{\mathrm{src}})$. The $\mathcal{I}^*$ will be defined more specifically below.

As $\mathcal{I}^*$  is produced with conditioning on $y_{\mathrm{src}}$, a proper way to obtain inner representations for the current trajectory, denoted as $\overline{\mathcal{I}}$, would be to make a single sampling step, starting at a current latent $z_t$ and conditioning on $y_{\mathrm{src}}$:

\vspace{-0.3cm}
\begin{equation}\label{eq:sampling_src}
\begin{array}{c}
     \bar{z}_{t-1} = a_t z_t + b_t \varepsilon_{\theta}(z_t, t, y_{\mathrm{src}}).
\end{array}
\end{equation}
\noindent Conditioning on the same prompt $y_{\mathrm{src}}$ is an important detail towards more stable editing process. This way, the resulting inner representations for the inversion and the current trajectories are aligned better, rather than when conditioning on different prompts. Besides, when preserving features of the initial image, we want to detach information about editing from the current state of these features. Our logic is, that with the sampling step in Equation \ref{eq:sampling_src} we want to see, how much has reconstruction of the initial image suffered, and compare it with the ideal reconstruction in Equation \ref{eq:sampling_rec}.

The current latent $z_t$ in Equation \ref{eq:sampling_src} is the result of applying CFG together with our self-guidance at the previous sampling step, initially being defined as $z_T = z^*_T$. 

We define the energy function as $g(z_t, z^*_t, t, y_{\mathrm{src}}, \mathcal{I}^*, \overline{\mathcal{I}})$. A sampling step in Equation \ref{eq:sampling_base} can be rewritten as:

\vspace{-0.3cm}
\begin{equation}\label{eq:rough_g}
    \begin{array}{c}
        \hat{\varepsilon}_{\theta}(z_t, t, y) = \mathrm{CFG}(z_t, t, y_{\mathrm{trg}}, 7.5) + v\cdot \nabla_{z_t} g(z_t, z^*_t, t, y_{\mathrm{src}}, \mathcal{I}^*, \overline{\mathcal{I}}).
    \end{array}
\end{equation}

See the overall pipeline of the method in Fig. \ref{fig:method_diagram} and more details can be found in Appendix \ref{app:pipeline}. We suggest ways to specify Equation \ref{eq:rough_g} by defining the inner representations sources in the following section, where we denote function $g$ as a \textit{guider}.

\subsubsection{Guiders.}
\label{sec:guiders}

We found that jointly including a Self-attention Guider and a Feature Guider into the generation process is sufficient for improving editing and preserving the initial image features. At the same time, they do not interfere with the editing, done by CFG. The effect of these guiders is illustrated in Fig \ref{fig:guiders_grid}. More details about guiders and its properties can be found in Appendix \ref{app:guiders}.

\noindent\textbf{Self-attention Guider.} As noted by the authors of \cite{Tumanyan_2023_CVPR}, self-attention maps contain information about the layout of the image, i.e. relative positioning of objects. While preserving layout is a primary challenge for stylisation tasks, it is also useful for other editing types.

To utilize this finding, we suggest guiding through matching of self-attention maps from the current trajectory $\bar{\mathcal{A}}^{\mathrm{self}}_i := \mathrm{self\;attn.}[\varepsilon_{\theta}(z_t, t, y_{\mathrm{src}})]$ and an ideal reconstruction trajectory $\mathcal{A}^{*\mathrm{self}}_i := \mathrm{self\;attn.}[\varepsilon_{\theta}(z^*_t, t, y_{\mathrm{src}})]$, where $i$ corresponds to the index of the UNet layer. 
So, in this case $\mathcal{I}^* = \{\mathcal{A}^{*\mathrm{self}}_i\}, \overline{\mathcal{I}} = \{\bar{\mathcal{A}}^{\mathrm{self}}_i\}$ and the guider is defined as follows:

\vspace{-0.4cm}
\begin{equation}\label{eq:self_attn}
    \begin{array}{c}
        g(z_t, z^*_t, t, y_{\mathrm{src}}, \{\mathcal{A}^{*\mathrm{self}}_i\}, \{\bar{\mathcal{A}}^{\mathrm{self}}_i\}) = \sum_{i = 1}^L\mathrm{mean}\|\mathcal{A}^{*\mathrm{self}}_i - \bar{\mathcal{A}}^{\mathrm{self}}_i \|^2_2,
    \end{array}
\end{equation}

\noindent where $L$ is a number of UNet layers.

\noindent\textbf{Feature Guider.} While preserving layout can significantly improve editing quality, it is not sufficient. For example, when applying local editing to a photo of a person, it is crucial to preserve visual features as well, such as face expressions.

To achieve this, for non-stylisation editing task we adopt an idea from \cite{epstein2023selfguidance}. The authors of this work propose an \textit{appearance} function for controlling visual appearance of an object, which utilizes masked \textit{features}. In this work we define \textit{features} $\Phi$ as an output of the last up-block in UNet. We also found, that masking is not crucial and does not influence the quality of editing in our setting, so we eliminated it for simplicity. Similar to \cite{epstein2023selfguidance}, we apply L1 norm to the difference of $\bar{\Phi} = \mathrm{features}[\varepsilon_{\theta}(z_t, t, y_{\mathrm{src}})]$ and $\Phi^* = \mathrm{features}[\varepsilon_{\theta}(z^*_t, t, y_{\mathrm{src}})]$:

\vspace{-0.3cm}
\begin{equation}\label{eq:features_other}
    \begin{array}{c}
        g(z_t, z^*_t, t, y_{\mathrm{src}}, \Phi^*, \bar{\Phi})
        = \mathrm{mean}\| \Phi^* - \bar{\Phi} \|_1.
    \end{array}
\end{equation}

For stylisation tasks, similar to \cite{Tumanyan_2023_CVPR}, we define $\Phi$ as an output of the second ResNet block in the second UNet up-block. Similar to Equation \ref{eq:features_other}, the guider is defined as follows:

\vspace{-0.3cm}
\begin{equation}\label{eq:features_stylization}
    \begin{array}{c}
        g(z_t, z^*_t, t, y_{\mathrm{src}}, \Phi^*, \bar{\Phi})
        = \mathrm{mean}\| \Phi^* - \bar{\Phi} \|^2_2.
    \end{array}
\end{equation}

\begin{figure}[h]
  \begin{subfigure}{.5\textwidth}
  \centering
    \includegraphics[width=.9\linewidth]{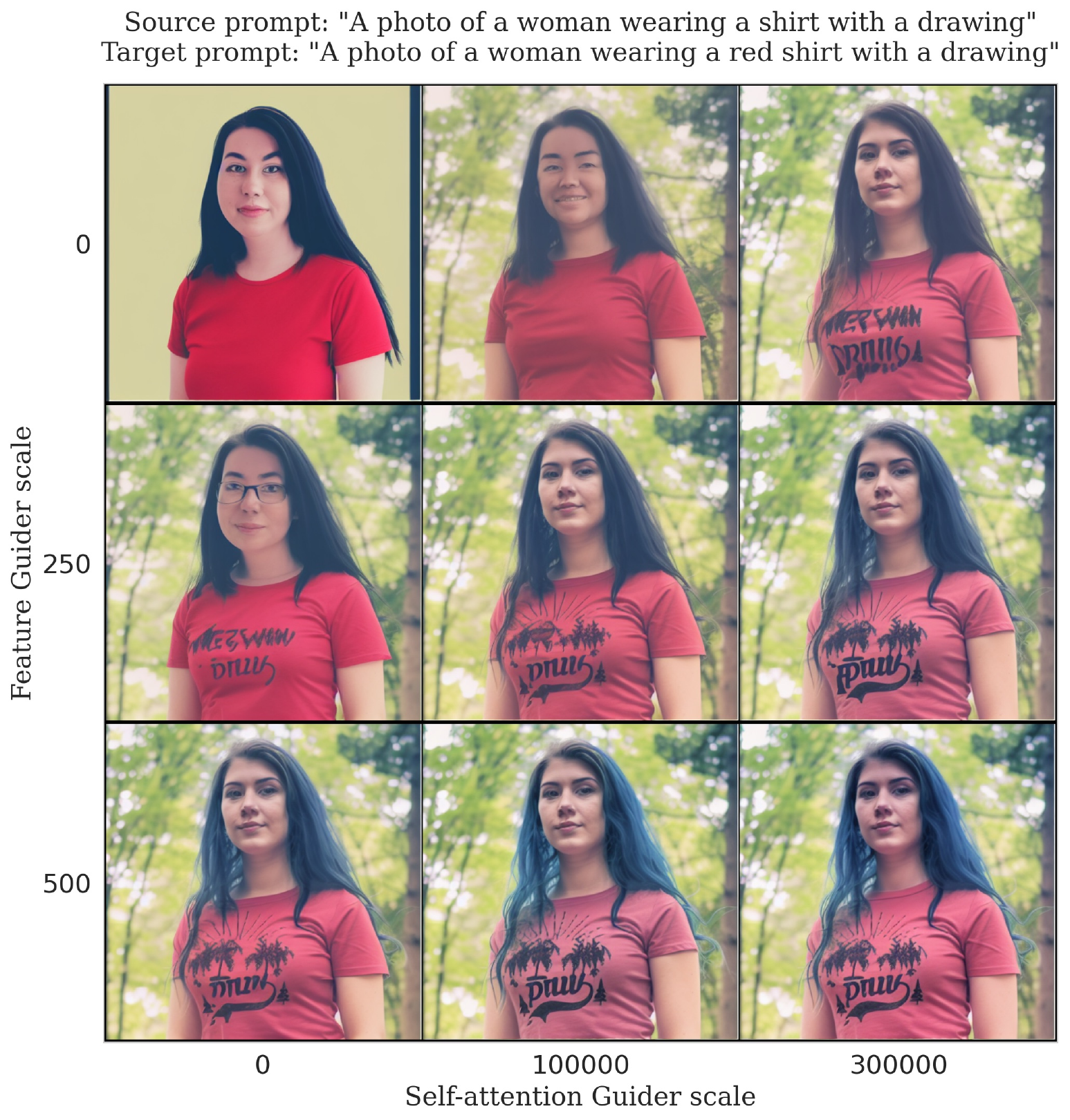}
    \caption{}
    \label{fig:guiders_grid}
  \end{subfigure}
  \begin{subfigure}{.5\textwidth}
  \centering
    \includegraphics[width=.9\linewidth]{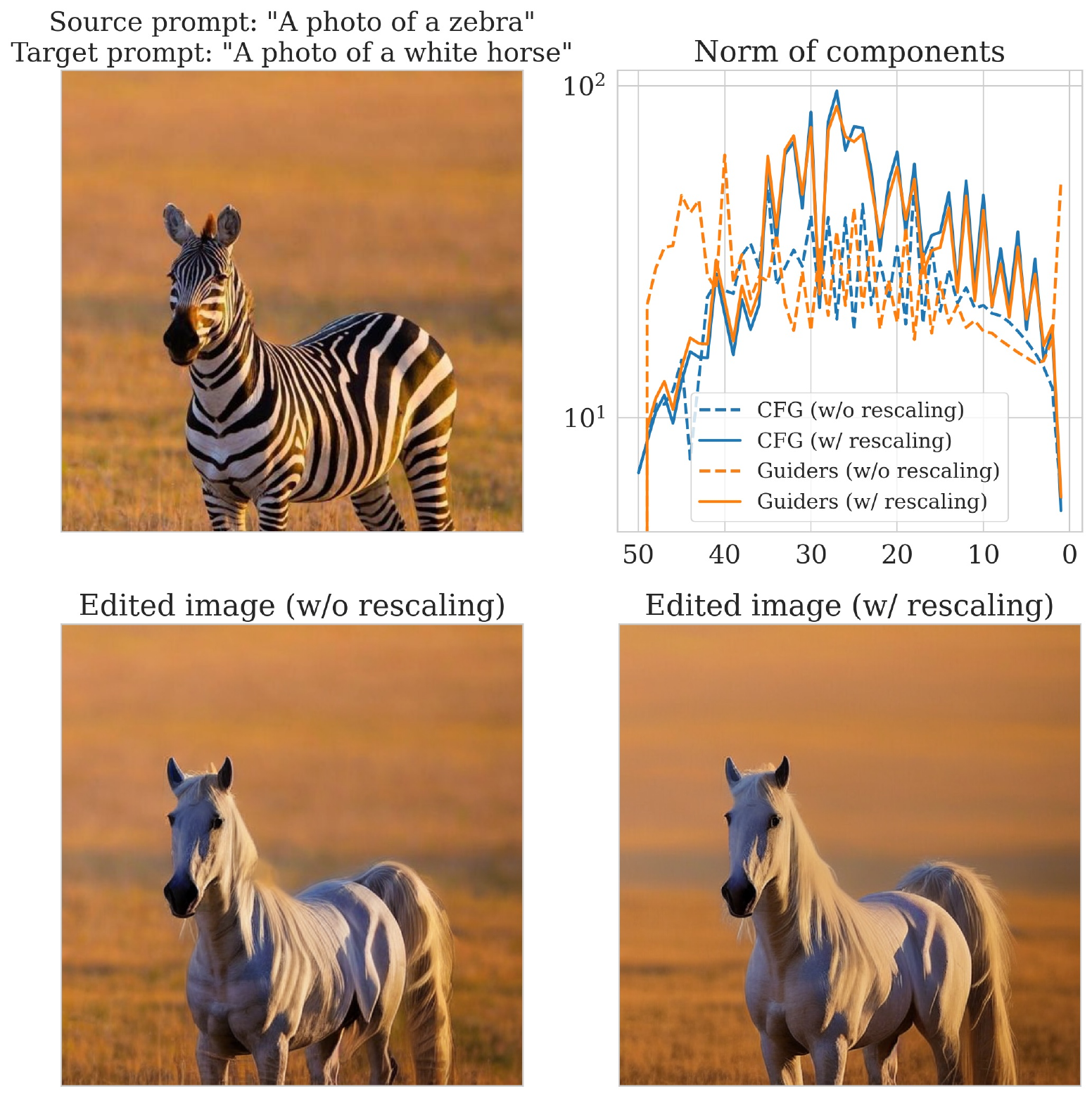}
    \caption{}
    \label{fig:norms_example}
  \end{subfigure}%
  \caption{\textbf{(a)} Effect of the proposed guiders (Equation \ref{eq:self_attn}, Equation \ref{eq:features_other}). Jointly applying both guiders preserves both layout and visual characteristics of unedited regions of the image. \textbf{(b)} Illustration of applying noise rescaling (Equation \ref{eq:noise_resc}). This technique aligns the sum of guiders with CFG according to the coefficient, defined in Equation \ref{eq:scaling_factor_def}, therefore stabilizes editing and improves its quality.}
\end{figure}

\subsubsection{Noise rescaling.}
\label{subsec:noise_rescaling}
Even though the proposed guiders significantly improve editing quality, the resulting method experiences inconsistency in terms of optimal guidance scales, when applied to a wide range of images.

We noticed that when our method's behavior is unstable, norms of CFG and the sum of gradients of all the guiders show instability over sampling steps as well.  

We suggest scaling the sum of guiders' gradients by a factor, depending on the CFG norm. This way, the proportion of both editing and preservation will be consistent throughout the whole synthesis process and easily controllable. 

More formally, a single noise sampling step will now be defined as:

\vspace{-0.3cm}
\begin{equation}\label{eq:noise_resc}
    \begin{array}{c}
        \epsilon_t = \varepsilon_{\theta}(z_t, t, \varnothing) + w \big( \varepsilon_{\theta}(z_t, t, y) \! - \! \varepsilon_{\theta}(z_t, t, \varnothing) \big) +\\
        + \gamma\sum_{i}v_i\cdot \nabla_{z_t} g_i(z_t, z^*_t, t, y_{\mathrm{src}}, \mathcal{I}^*, \overline{\mathcal{I}}),
    \end{array}
\end{equation}

\noindent where $\gamma$ is a scaling factor. With $r$ being a numerical parameter, controlling the proportion between editing and preservation, $\gamma$ can be computed as:

\vspace{-0.3cm}
\begin{equation}\label{eq:scaling_factor_def}
    \begin{array}{c}
         r_{\mathrm{cur}}(t) = {\displaystyle\frac{\|w \big( \varepsilon_{\theta}(z_t, t, y) - \varepsilon_{\theta}(z_t, t, \varnothing) \big)\|^2_2}{\|\sum_{i}v_i\cdot \nabla_{z_t} g_i(z_t, z^*_t, t, y_{\mathrm{src}})\|^2_2}},\\\\
         \gamma = r \cdot r_{\mathrm{cur}}(t).
    \end{array}
\end{equation}

We found that keeping $r$ in an interval, rather than defining it as a fixed numerical hyperparameter, works best in experiments. An interval can be defined so that the fraction of norms in Equation \ref{eq:scaling_factor_def} does not exceed some pre-defined upper and lower boundaries. This way, together with obtaining the desired consistency, we still let the diffusion model decide, whether the current sampling step needs more editing or preservation. More formally:

\vspace{-0.3cm}
\begin{equation}\label{eq:scaling_in_range}
    r(t) = \left\{ \begin{array}{ll}
        r_{\mathrm{lower}}, &  \frac{1}{r_{\mathrm{cur}}(t)} \le r_{\mathrm{lower}} \\
         \frac{1}{r_{\mathrm{cur}}(t)}, & r_{\mathrm{lower}} <  \frac{1}{r_{\mathrm{cur}}(t)} < r_{\mathrm{upper}}\\
         r_{\mathrm{upper}}, & \frac{1}{r_{\mathrm{cur}}(t)} \ge r_{\mathrm{upper}}
    \end{array} \right.,
\end{equation}

\noindent An example of applying noise rescaling to the proposed method is provided in Fig. \ref{fig:norms_example}. More details about rescaling and its analysis can be found in Appendix \ref{app:rescaling}.

\section{Experiments}
\label{sec:experiments}

\textbf{Experiment Setup.}
In order to comprehensively compare image manipulation methods, we decided to consider 4 different types of editing: 1) local manipulation of a person's appearance and clothing, 2) changing person's emotions, 3) replacing one animal with another, and 4) global image stylisation. For each type, we collected 20 different example edits on which we tested all methods. A more detailed description of these types of edits and all examples can be found in the Appendix \ref{app:dataset}. 

To compare the similarity of edited and real images we collected more extensive datasets. A particular case of animal-to-animal type of editing is the task of transformation dogs into cats. To compare methods on this task we collected 500 examples of dogs from the AFHQ dataset \cite{choi2020starganv2} and then compared the results of editing with cats from the same dataset. For other types of editing datasets and details are described in the Appendix \ref{app:dataset}.

We compared our method with existing approaches that have publicly available source code: NPI \cite{miyake2023negativeprompt}, NPI Prox \cite{han2023improving} and NTI \cite{mokady2022null} based on P2P \cite{hertz2022prompt}, MasaCtrl \cite{cao_2023_masactrl}, ProxMasaCtrl \cite{han2023improving}, PnP \cite{Tumanyan_2023_CVPR}, and EDICT \cite{10204740}. We used the authors' original code with the default parameters recommended in the description of each method. For our method, we also fixed hyperparameters during evaluation. More details about our method setup can be found in Appendix \ref{app:setup}.

\begin{figure*} 
    \includegraphics[width=\linewidth]{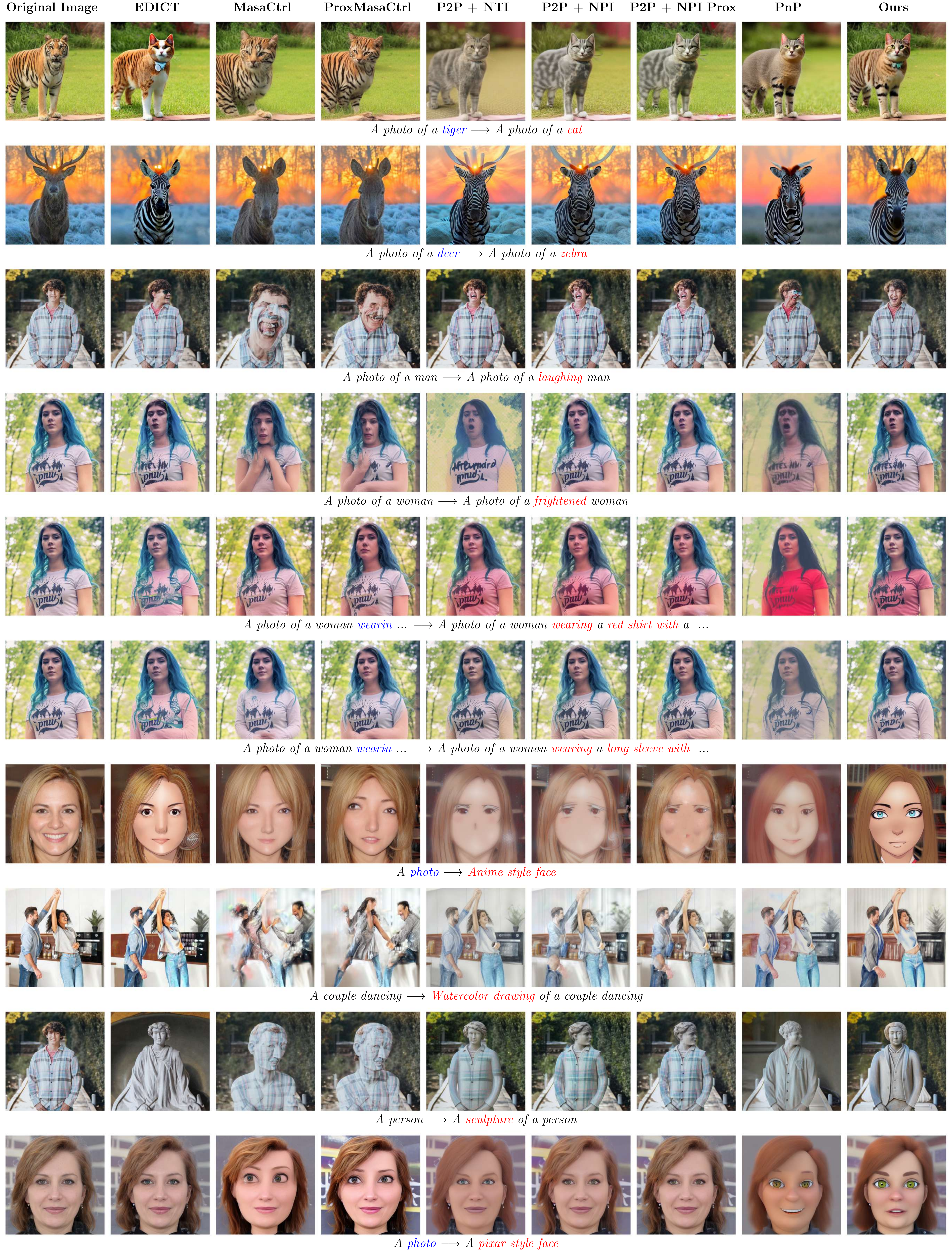}
    \caption{Visual comparison of our method with baselines over different types of editing. Our approach shows more consistent results than existing methods and achieves a better trade-off between editing quality and preservation of the structure of the original image.}
    \label{fig:visual_comparison}
\end{figure*}

\subsection{Qualitative Comparison}
In Fig. \ref{fig:visual_comparison} we see the result of our method and previous approaches on examples of edits belonging to one of the 4 types of edits mentioned above. It can be seen that our method shows stable quality across all types of edits. In cases where the whole image does not need to be stylized, our method, unlike baselines, is good at preserving the structure and background that should not be changed. For example, when editing a person's emotion, our method only affects the face area and does not change the rest of the person's attributes and background, unlike other methods that corrupt the overall structure of the original image. This can be seen particularly well in local edits of different attributes of a woman. The MasaCtrl and ProxMasaCtrl methods strongly alter the geometry of the person, and the target edit is not always obtained. The P2P+NTI, P2P+NPI, and P2P+NPI Prox methods either transform the target attribute while corrupting others or edit nothing at all. The PnP method almost always achieves the desired edit but at the cost of severe loss of structure of the original image. The EDICT method performs best compared to the other methods but is still inferior to our method in most of these examples.

\subsection{Quantitative Comparison}
To quantitatively evaluate the performance of our method, we decided to use an evaluation protocol similar to that used in EDICT \cite{10204740}. This protocol measures two key properties of the method. The first is the extent to which the result of the method contains the desired edit. To measure this characteristic, we use the CLIP score \cite{radford2021learning}, which is calculated between the edited image and the target prompt. The second property is the extent to which the method preserves the structure and details of the original image that should not change during editing. For this, we use the LPIPS metric \cite{zhang2018unreasonable} between the original image and its edited version. 

The results of these metrics for our method and others are shown in Table \ref{tab:quant_comparison}. They show that our method achieves the best balance between preserving the original image structure and editing the target attribute. In particular, on the LPIPS metric, we perform best apart from the EDICT method, which we significantly outperform on the CLIP score. On the CLIP score, we also perform best apart from the PnP method, which is strongly inferior in its ability to preserve the structure of the original image in terms of the LPIPS metric. The other approaches show uniformly worse results with respect to these two metrics. It is also worth noting that compared to the most advanced EDICT and PnP methods, our method is more computationally efficient, as can be seen from the running time for editing an image. More analysis of these results can be found in Appendix \ref{app:quant}.

To quantitatively evaluate the distance between the distributions of edited and real images we calculate FID Score \cite{Seitzer2020FID} on the dog-to-cat transformation task. As shown in Table \ref{tab:quant_comparison} our method outperforms the others by FID Score. MasaCtrl and ProxMasaCtrl show the highest FID that is correlated with artifacts on edited images and its nonrealism in Fig. \ref{fig:visual_comparison_dog_to_cat} in Appendix \ref{app:afhq_description}. The closest FID Score to our method has PnP but it is shown to be several times slower than our method. 

\begin{table}[h!]
    \vspace{-0.2cm}
    \centering
    \caption{Comparison with baselines on image editing task by using 80 examples of 4 different edit types to compute the LPIPS and CLIP per edit. 500 samples of the dog-to-cat type of editing are used to compute FID. Our method achieves the best balance between editing quality (CLIP) and preservation of the original image (LPIPS) and reaches the best similarity to the real image's distribution (FID). The inference time (including the inversion part) is computed on a single GPU A100 in seconds.}
    \small
    \begin{tabular}{lcccc}
    \toprule
        \textbf{\quad \ \ \ Method}         & LPIPS $\downarrow$ & CLIP $\uparrow$ & FID $\downarrow$ & Time (s) $\downarrow$ \\
        \midrule
        
ProxMasaCtrl \cite{han2023improving} &  0.267 &  0.215 &  94.53 & 12.94 \\
MasaCtrl \cite{cao_2023_masactrl} &  0.306 &  0.223 &  100.62 & 13.73 \\
EDICT \cite{10204740} &  \underline{0.221} &  0.229 &  47.13 & 68.13 \\
P2P \cite{hertz2022prompt} + NTI \cite{mokady2022null} &  0.279 &  0.233 &  42.46 & 66.77 \\
P2P \cite{hertz2022prompt} + NPI \cite{miyake2023negativeprompt} Prox \cite{han2023improving} &  \textbf{0.170} &  0.233 &  43.16 & 8.59 \\
P2P \cite{hertz2022prompt} + NPI \cite{miyake2023negativeprompt} &  0.251 &  0.234 &  44.05 & 8.54 \\
PnP \cite{Tumanyan_2023_CVPR} &  0.366 &  \textbf{0.256} &  \underline{39.55} & 197.0 \\
\midrule
Guide-and-Rescale (ours) &  0.228 &  \underline{0.243} & \textbf{39.07} & 24.26 \\
        
        \bottomrule
    \end{tabular}
    \label{tab:quant_comparison}
    \vspace{-0.5cm}
\end{table}

\subsection{User Study}
\label{sec:experiments_sub_userstudies}

Next, we conducted a user study to compare our method with baselines. We used the same example edits of the 4 types we described above. We evaluate how much our method is preferred by others in the form of a pairwise comparison question. We surveyed 62 users who answered a total of 960 questions. Each question showed the original image, the targeting prompt, and two results. Users had to answer two questions. The first question (Q1) is "Which image matches the desired editing description best?". The second question (Q2) is "Which image preserves the overall structure of the "Original Image" best?".

The results of this user study are shown in Table \ref{tab:user_study}. Each value represents the percentage of the users that preferred our method over the corresponding baseline in the first column. We can see that our method is preferred over almost all baselines in terms of "Editing" quality and is subjectively similar to P2P + NPI Prox. However, our method is better than P2P + NPI Prox in terms of preservation quality. To summarize, our method is consistent over a wide range of different edits and subjectively gives better results on editing quality while preserving source image regions that should not be edited. Details of the user study can be found in the Appendix \ref{app:user_study}.

\begin{table}[h!]
    \vspace{-0.2cm}
    \centering
    \caption{User study that evaluates the subjective preference of our method that is preferred over all baselines.}
    \small
    \begin{tabular}{lccc}
    \toprule
        \textbf{\quad \ \ \ Method} & Editing (Q1) & Preservation (Q2) \\
        \midrule
MasaCtrl \cite{cao_2023_masactrl} & 85 \% & 70 \% \\
ProxMasaCtrl \cite{han2023improving} & 82 \% & 63 \% \\
P2P \cite{hertz2022prompt} + NTI \cite{mokady2022null} & 60 \% & 49 \% \\
P2P \cite{hertz2022prompt} + NPI \cite{miyake2023negativeprompt} & 59 \% & 58 \% \\
P2P \cite{hertz2022prompt} + NPI \cite{miyake2023negativeprompt} Prox \cite{han2023improving} & 50 \% & 55 \% \\
PnP \cite{Tumanyan_2023_CVPR} & 60 \% & 61 \% \\
EDICT \cite{10204740} & 56 \% & 59 \% \\
        \bottomrule
    \end{tabular}
    \label{tab:user_study}
    \vspace{-0.5cm}
\end{table}

\section{Conclusion}
In this paper, we propose a novel image editing method Guide-and-Rescale based on the self-guidance mechanism and show its effectiveness for a variety of editing types. It significantly improves the trade-off between editing quality and original image preservation. We show that it achieves more consistent and high-quality results than existing approaches both qualitatively and quantitatively.

\par\vfill\par

\section*{Acknowledgements}
The analysis of related work in sections 1 and 2 were obtained by Aibek Alanov with the support of the grant for research centres in the field of AI provided by the Analytical Center for the Government of the Russian Federation (ACRF) in accordance with the agreement on the provision of subsidies (identifier of the agreement 000000D730321P5Q0002) and the agreement with HSE University No. 70-2021-00139.
This research was supported in part through computational resources of HPC facilities at HSE University.

%
%
\bibliographystyle{splncs04}
\bibliography{main}

\newpage
\appendix
\onecolumn


\begin{center}
    \textbf{\Large Appendix}
\end{center}

\section{Overall Pipeline}
\label{app:pipeline}

The proposed method is summarized in Alg. \ref{alg:overall_pipeline}.

We suggest setting the CFG scale equal standard sampling guidance scale for Stable Diffusion $w = 7.5$. For non-stylisation tasks, we set Self-attention Guider scale $v_{\mathrm{self}} = 300000$ and Feature Guider scale $v_{\mathrm{feat}} = 500$, while for stylisation editing optimal guidance scales equal $v_{\mathrm{self}} = 100000, v_{\mathrm{feat}} = 2.5$. Besides, we suggest disabling guiders for several last steps of synthesis for more stable editing. For this purpose, we define a threshold $\tau$, denoting the number of synthesis steps when guiders are used. For non-stylisation tasks $\tau = 35$, and for stylisation: $\tau = 25$.

Noise rescaling boundaries for non-stylisation are suggested as $r_{\mathrm{lower}} = 0.33$, $r_{\mathrm{upper}} = 3$, while for stylisation task we recommend $r_{\mathrm{fixed}} := r_{\mathrm{upper}} = r_{\mathrm{lower}} = 1.5$. 

Regarding notations in Alg. \ref{alg:overall_pipeline}, DDIM  sampling formula takes form of:
\begin{equation}\label{eq:ddim_sampling_appendix}
    \mathrm{DDIM\;Sample}(z_t, \epsilon_t) = a_tz_t + b_t \epsilon_t.
\end{equation}
\noindent DDIM inversion formula stays unchanged, as defined in Equation \ref{eq:ddim_inversion}. It is also important to note, that coefficients used in Equation \ref{eq:ddim_sampling_appendix} and Equation \ref{eq:ddim_inversion} are defined as follows:

\vspace{-0.3cm}
\begin{equation}
    \begin{array}{c}
        a_t = \sqrt{\alpha_{t-1}/\alpha_t}, \\
        b_t = \sqrt{\alpha_{t-1}}\left( \sqrt{1/\alpha_{t-1} - 1} - \sqrt{1/\alpha_t - 1}\right),\\\\
        a^*_t = \sqrt{\alpha_{t + 1}/\alpha_{t}},\\
        b^*_t = \sqrt{\alpha_{t + 1}}\left( \sqrt{1/\alpha_{t + 1} - 1} - \sqrt{1/\alpha_{t} - 1}\right).
    \end{array}
\end{equation}
\noindent We use DDIM with $T = 50$ inner steps.

For simplicity, we eliminate dependency on $z_t, z^*_t, t, y_{\mathrm{src}}$ from guiders in Equation \ref{eq:self_attn} and Equations \ref{eq:features_other}, \ref{eq:features_stylization} and obtain $g_{\mathrm{self}}$ and $g_{\mathrm{feat}}$ correspondingly.

Besides, we omit dependency on current timestep $t$ from noise rescaling definition in Equation \ref{eq:scaling_in_range} and reformulate it as:

\vspace{-0.3cm}
\begin{equation}\label{eq:scaling_in_range_appendix}
    f_{\gamma}(r_{\mathrm{l}}, r_{\mathrm{u}}, r_{\mathrm{cur}}) = \left\{ \begin{array}{ll}
        r_{\mathrm{l}} \cdot r_{\mathrm{cur}}, &  1/r_{\mathrm{cur}} \le r_{\mathrm{l}} \\
         1, & r_{\mathrm{l}} < 1/ r_{\mathrm{cur}} < r_{\mathrm{u}}\\
         r_{\mathrm{u}} \cdot r_{\mathrm{cur}}, & 1/r_{\mathrm{cur}} \ge r_{\mathrm{u}}
    \end{array} \right. .
\end{equation}

\begin{algorithm}[t]
\caption{Guide-and-Rescale for Real Image Editing}\label{alg:overall_pipeline}
    \begin{algorithmic}[1]
        \Input Real image $x_{\mathrm{init}}$, source prompt $y_{\mathrm{src}}$, target prompt $y_{\mathrm{trg}}$; DDIM steps $T$; guidance scales $w$, $v_{\mathrm{self}}$, $v_{\mathrm{feat}}$; threshold $\tau$; noise rescaling boundaries $r_{\mathrm{lower}}$, $r_{\mathrm{upper}}$.
        \Function VAE encoder $Enc.$, VAE decoder $Dec.$, $\mathrm{DDIM\;Inversion}$ (Equation \ref{eq:ddim_inversion}), $\mathrm{DDIM\;Sample}$ (Equation \ref{eq:ddim_sampling_appendix}), Self-attention Guider $g_{\mathrm{self}}$ (Equation \ref{eq:self_attn}), Feature Guider $g_{\mathrm{feat}}$ (Equations \ref{eq:features_other}, \ref{eq:features_stylization}), noise rescaling $f_{\gamma}$ (Equation \ref{eq:scaling_in_range_appendix}).
        \Output Edited image $x_{\mathrm{edit}}$.
        \\$z^*_0 = Enc.(x_{\mathrm{init}})$
        \For{$t=0,1,\ldots,T-1$}
            \State $z^*_{t+1} = \mathrm{DDIM\;Inversion}(z^*_t, y_{\mathrm{src}})$
        \EndFor
        \\$z_T = z^*_T$
        \For{$t= T,T-1,\ldots,1$}
            \State $\Delta_{\mathrm{cfg}} = w (\varepsilon_{\theta}(z_t, t, y_{\mathrm{trg}}) - \varepsilon_{\theta}(z_t, t,\varnothing))$
            \State $\epsilon_{\mathrm{cfg}} = \varepsilon_{\theta}(z_t, t, \varnothing) + \Delta_{\mathrm{cfg}}$
            \State $\big\{ \{\mathcal{A}^{*\mathrm{self}}_i\}_{i=1}^L, \Phi^* \big\}= \varepsilon_{\theta}(z^*_t, t, y_{\mathrm{src}}) $
            \State $\big\{ \{\bar{\mathcal{A}}^{\mathrm{self}}_i\}_{i=1}^L, \bar{\Phi} \big\}= \varepsilon_{\theta}(z_t, t, y_{\mathrm{src}}) $
            \State $\epsilon_{\mathrm{self}} = v_{\mathrm{self}} \cdot g_{\mathrm{self}}(\{\mathcal{A}^{*\mathrm{self}}_i\}_{i=1}^L, \{\bar{\mathcal{A}}^{\mathrm{self}}_i\}_{i=1}^L)$
            \State $\epsilon_{\mathrm{feat}} = v_{\mathrm{feat}} \cdot g_{\mathrm{feat}}(\Phi^*, \bar{\Phi})$
            \State $r_{\mathrm{cur}} = \|\Delta_{\mathrm{cfg}}\|^2_2 / \| \nabla_{z_t} (\epsilon_{\mathrm{self}} +\epsilon_{\mathrm{feat}}) \|^2_2$
            \State $\gamma = f_{\gamma}(r_{\mathrm{lower}}, r_{\mathrm{upper}}, r_{\mathrm{cur}})$
            \If{$T - t < \tau$}
                \State $\epsilon_{\mathrm{final}} = \epsilon_{\mathrm{cfg}} + \gamma \cdot \nabla_{z_t} (\epsilon_{\mathrm{self}} +\epsilon_{\mathrm{feat}})$
            \Else
                \State $\epsilon_{\mathrm{final}} = \epsilon_{\mathrm{cfg}}$
            \EndIf
            \State$z_{t-1} = \mathrm{DDIM\;Sample}(z_t, \epsilon_{\mathrm{final}})$
        \EndFor
        \\$x_{\mathrm{edit}} = Dec.(z_0)$
        \Return $x_{\mathrm{edit}}$
    \end{algorithmic}
\end{algorithm}

\section{Guiders}
\label{app:guiders}

In Sec. \ref{sec:method_guide_and_resacle} we propose Self-attention Guider and Feature Guider.

\begin{figure*}[t]
    \centering
    \begin{subfigure}{\textwidth}
    	\includegraphics[width=\linewidth]{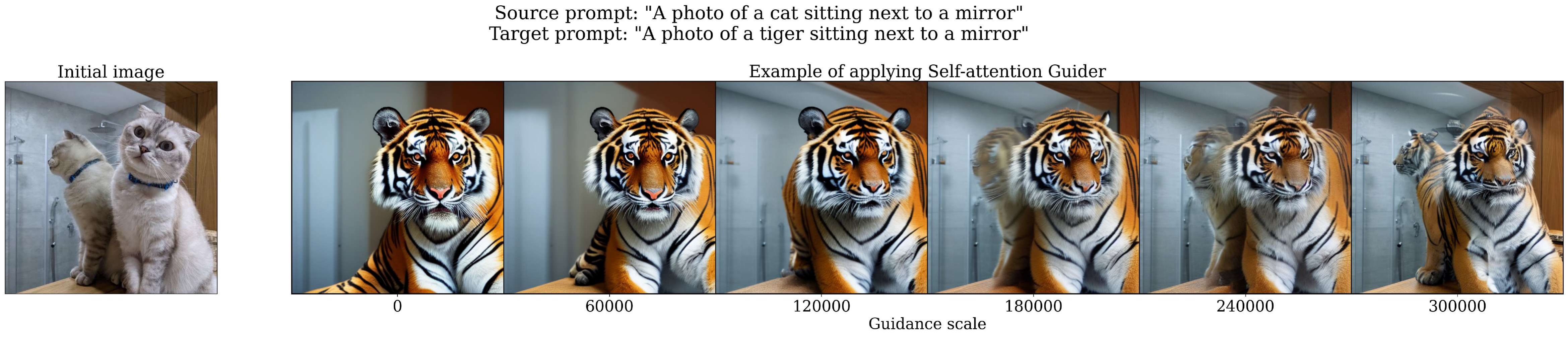}
    	\caption{Non-stylisation editing example.}
        \label{fig:self_attn_guidance_scale_non_styl}
    \end{subfigure}
    \par\bigskip
    \begin{subfigure}{\textwidth}
        \includegraphics[width=\linewidth]{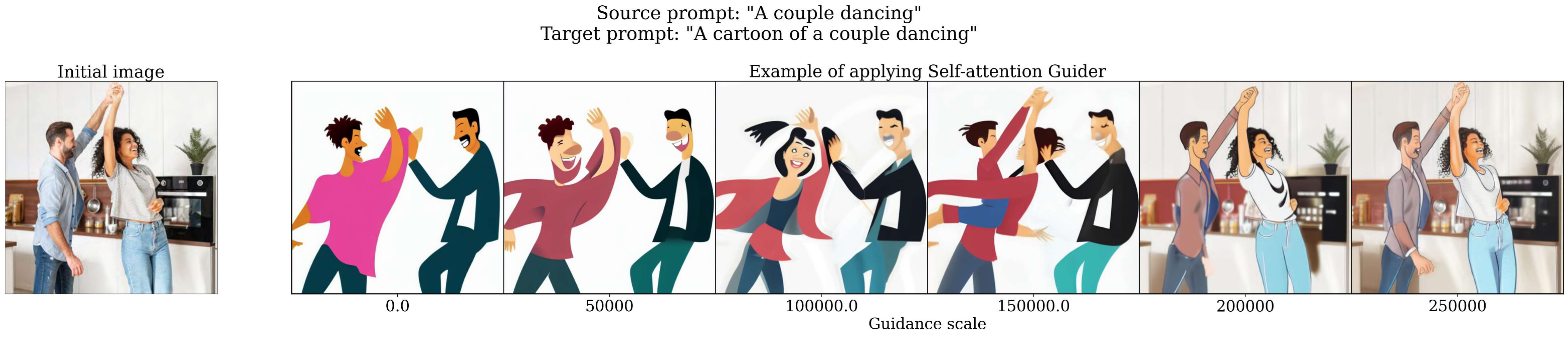}
    	\caption{Stylisation editing example.}
        \label{fig:self_attn_guidance_scale_styl}
    \end{subfigure}
    \vspace{-0.5cm}
    \caption{Illustration of applying \textbf{Self-attention Guider}. Self-attention Guider is defined in Equation \ref{eq:self_attn}.}
    \label{fig:self_attn_guidance_scale}
\end{figure*}

\noindent \textbf{Self-attention Guider.} In Fig. \ref{fig:self_attn_guidance_scale} we show examples of applying this guider. We use this guider alone with CFG, and visualize the results of editing with different guidance scales $v$:

\vspace{-0.3cm}
\begin{equation}
    \begin{array}{c}
        \epsilon_{\mathrm{final}} = \mathrm{CFG}(z_t, t, y_{\mathrm{trg}}, 7.5) + v \cdot \nabla g_{\mathrm{self}}( \{\mathcal{A}_i^{*\mathrm{self}}\}_{i=1}^L, \{\bar{\mathcal{A}}_i^{\mathrm{self}}\}_{i=1}^L ).
    \end{array}
\end{equation}

This guider is formally defined in Equation \ref{eq:self_attn}. We calculate L2 norm of the difference of self-attention maps from the current sampling trajectory, conditioning on $y_{\mathrm{src}}$, and the reference inversion trajectory, where by L2 norm we mean the following:

\vspace{-0.3cm}
\begin{equation}\label{eq:l2_norm_def}
    \|X-Y\|^2_2 = (X - Y)^2,
\end{equation}
\noindent where arithmetic operations are applied element-wise to the matrices of the same shape.

We take the mean of the resulting matrix and then sum these results over all UNet layers.

\begin{figure*}[t]
    \centering
    \begin{subfigure}{\textwidth}
    	\includegraphics[width=\linewidth]{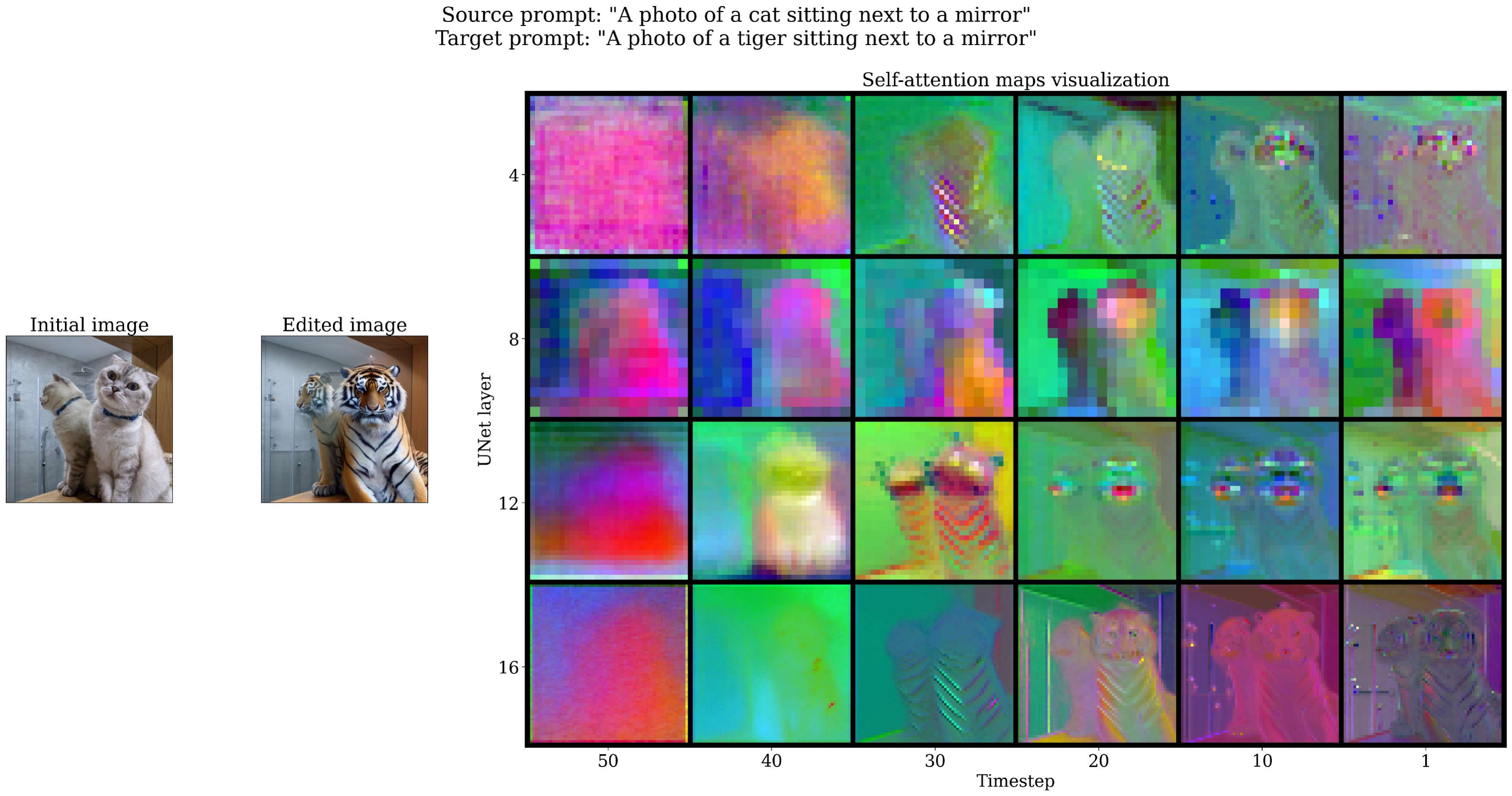}
    	\caption{Self-attention maps $\{\bar{\mathcal{A}}_i^{\mathrm{self}}\}$ from the current trajectory $\varepsilon_{\theta}(z_t, t, y_{\mathrm{src}})$}
        \label{fig:self_attn_guider_visualization_non_styl_cur_inv}
    \end{subfigure}
    \par\bigskip
    \begin{subfigure}{\textwidth}
    	\includegraphics[width=\linewidth]{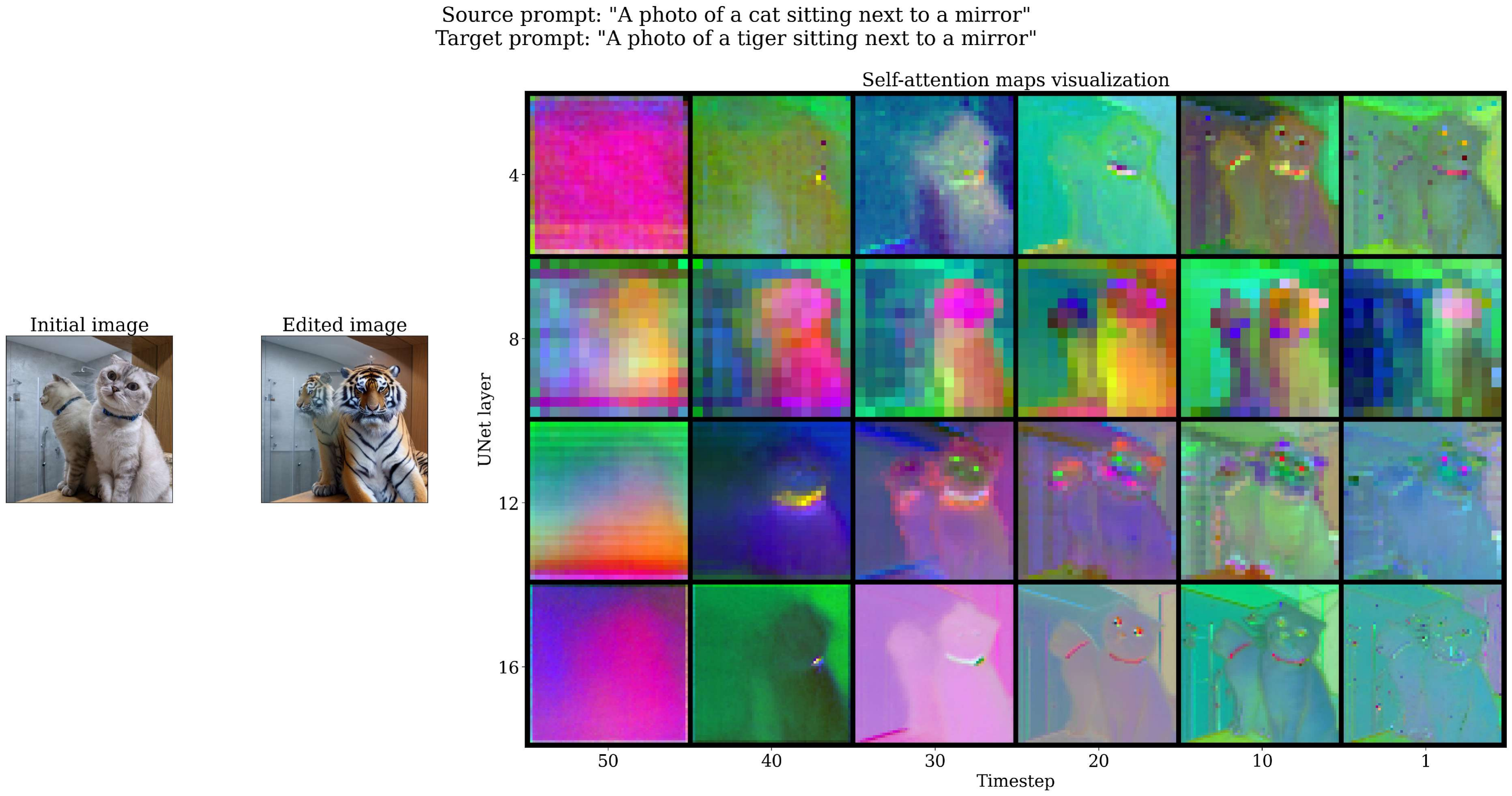}
    	\caption{Self-attention maps $\{\mathcal{A}_i^{*\mathrm{self}}\}$ from the reference inversion trajectory $\varepsilon_{\theta}(z^*_t, t, y_{\mathrm{src}})$}
        \label{fig:self_attn_guider_visualization_non_styl_inv_inv}
    \end{subfigure}
    \vspace{-0.5cm}
    \caption{Visualization of self-attention maps from Self-attention Guider (Equation \ref{eq:self_attn}) for non-stylisation editing example.}
    \label{fig:self_attn_guider_visualization_non_styl}
\end{figure*}

\begin{figure*}[t]
    \centering
    \begin{subfigure}{\textwidth}
    	\includegraphics[width=\linewidth]{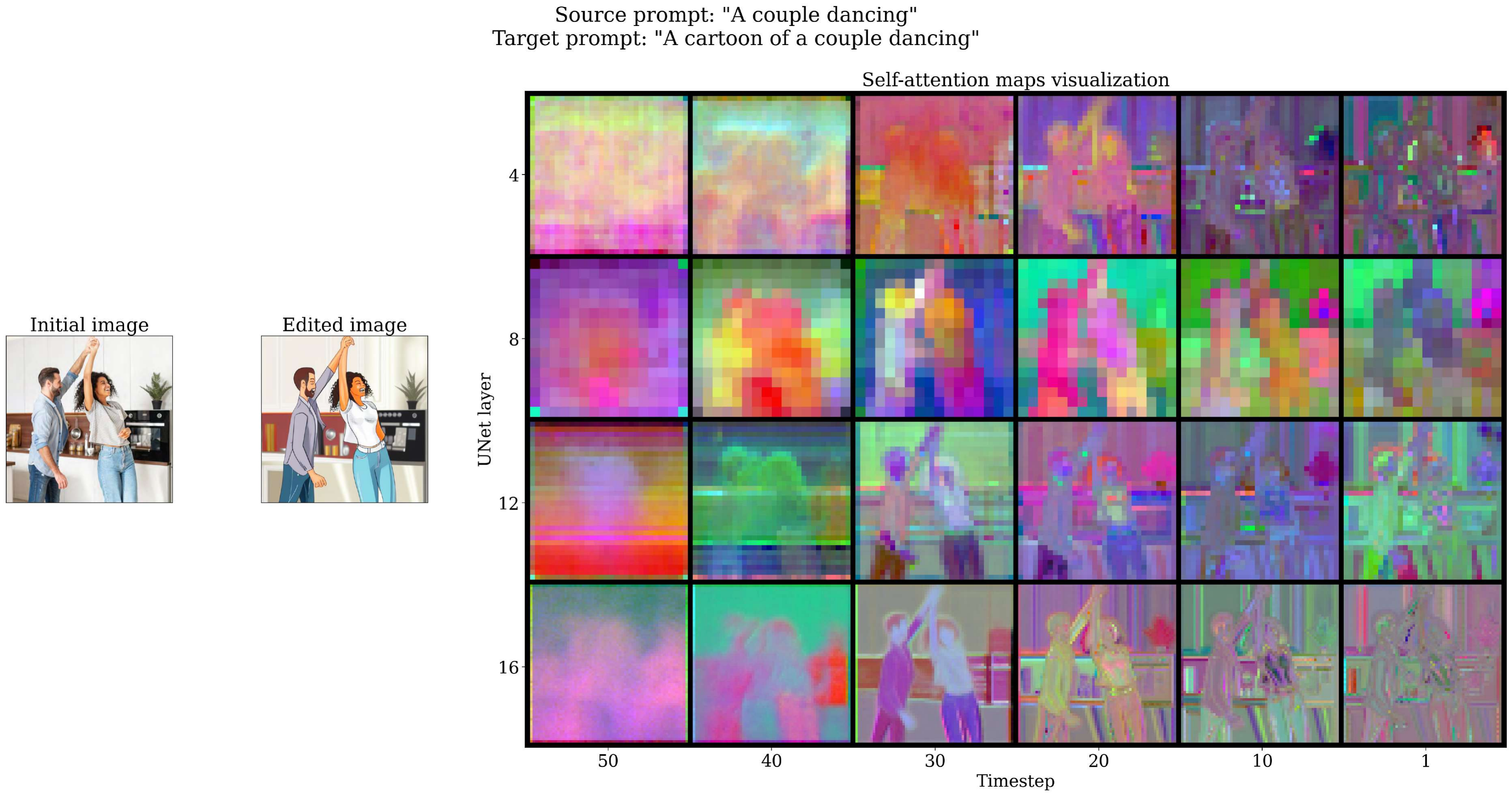}
    	\caption{Self-attention maps $\{\bar{\mathcal{A}}_i^{\mathrm{self}}\}$ from the current trajectory $\varepsilon_{\theta}(z_t, t, y_{\mathrm{src}})$}
        \label{fig:self_attn_guider_visualization_styl_cur_inv}
    \end{subfigure}
    \par\bigskip
    \begin{subfigure}{\textwidth}
    	\includegraphics[width=\linewidth]{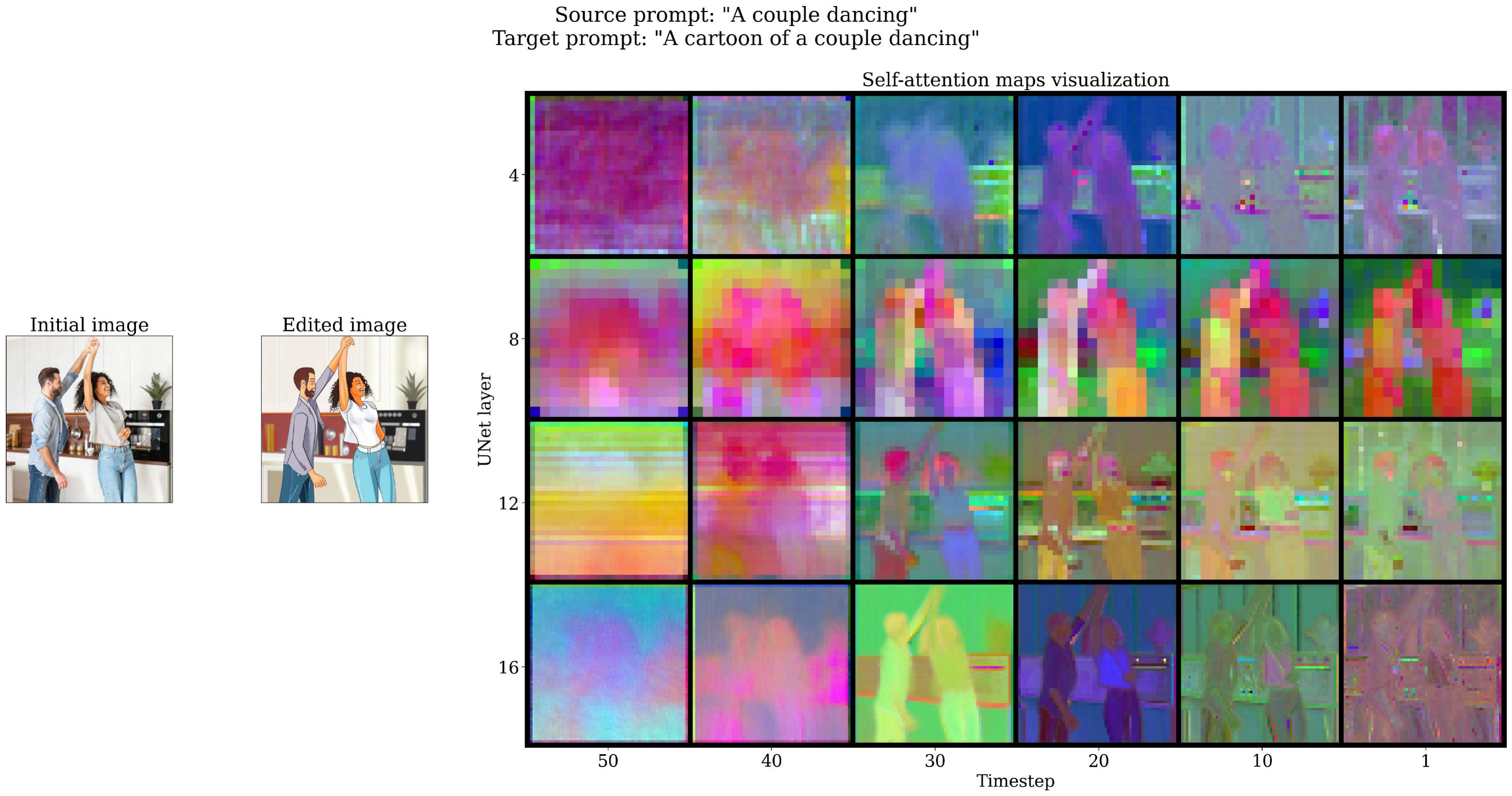}
    	\caption{Self-attention maps $\{\mathcal{A}_i^{*\mathrm{self}}\}$ from the reference inversion trajectory $\varepsilon_{\theta}(z^*_t, t, y_{\mathrm{src}})$}
        \label{fig:self_attn_guider_visualization_styl_inv_inv}
    \end{subfigure}
    \vspace{-0.5cm}
    \caption{Visualization of self-attention maps from Self-attention Guider (Equation \ref{eq:self_attn}) for stylisation editing example.}
    \label{fig:self_attn_guider_visualization_styl}
\end{figure*}

This guider preserves image layout, i.e. relative positioning of objects. In Fig. \ref{fig:self_attn_guider_visualization_non_styl} and Fig. \ref{fig:self_attn_guider_visualization_styl} we prove this statement by visualizing three leading principal components of self-attention maps. The visualization shows, that the colors are aligned with objects in the picture.

\begin{figure*}[t]
    \centering
    \begin{subfigure}{\textwidth}
    	\includegraphics[width=\linewidth]{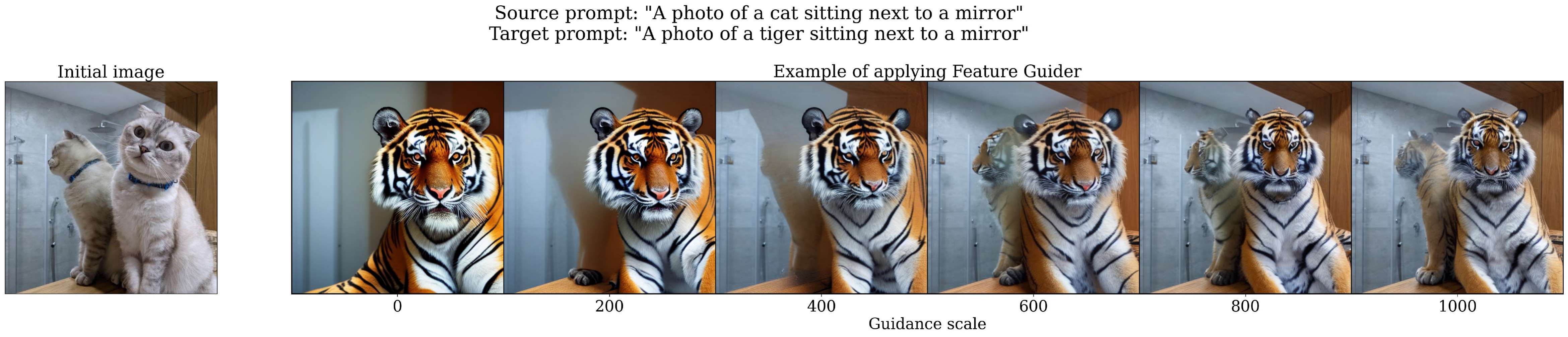}
    	\caption{Non-stylisation editing example. Feature Guider is defined in Equation \ref{eq:features_other}.}
        \label{fig:feature_guidance_scale_non_styl}
    \end{subfigure}
    \par\bigskip
    \begin{subfigure}{\textwidth}
        \includegraphics[width=\linewidth]{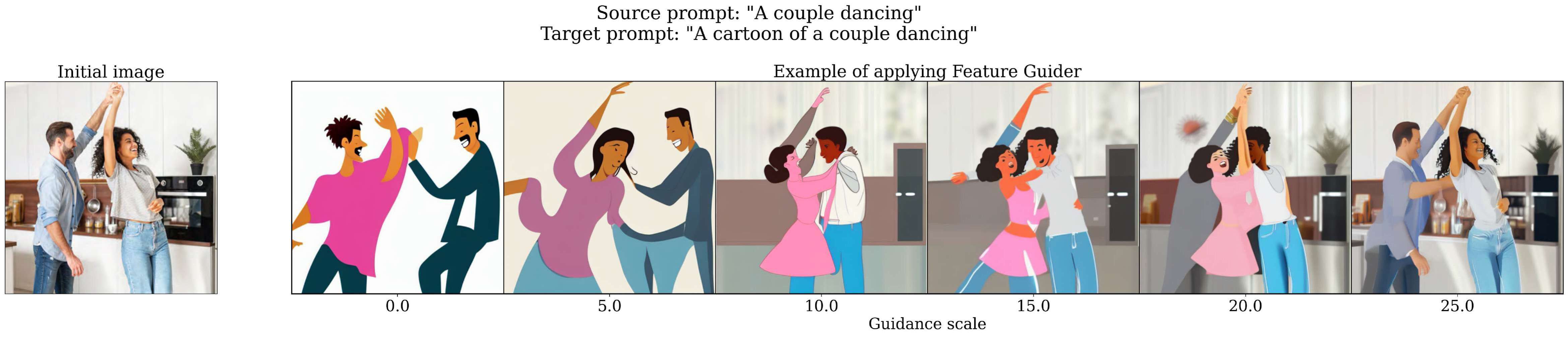}
    	\caption{Stylisation editing example. Feature Guider is defined in Equation \ref{eq:features_stylization}.}
        \label{fig:feature_guidance_scale_styl}
    \end{subfigure}
    \vspace{-0.5cm}
    \caption{Illustration of applying \textbf{Feature Guider}.}
    \label{fig:feature_guidance_scale}
\end{figure*}

\begin{figure*}[t]
    \centering
    \begin{subfigure}{\textwidth}
    	\includegraphics[width=\linewidth]{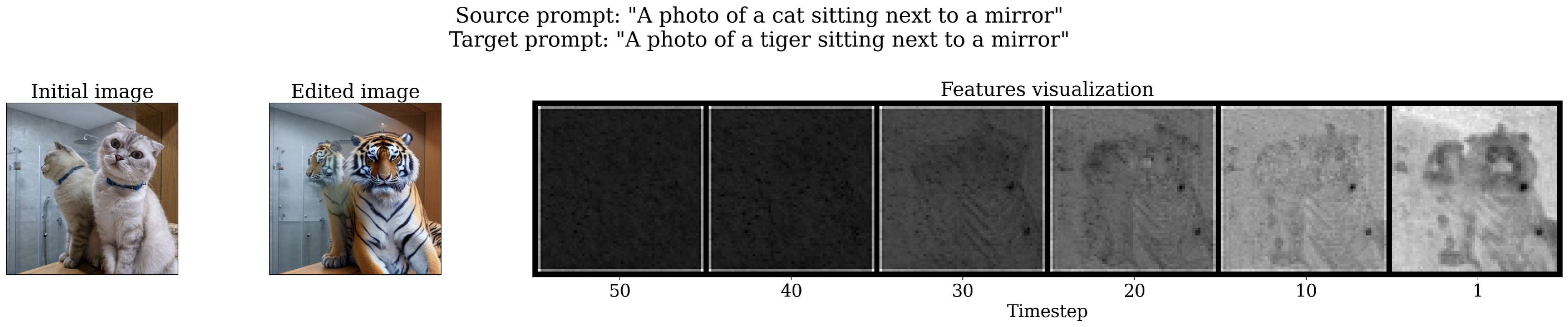}
    	\caption{Features $\bar{\Phi}$ from the current trajectory $\varepsilon_{\theta}(z_t, t, y_{\mathrm{src}})$}
        \label{fig:feature_guider_visualization_non_styl_cur_inv}
    \end{subfigure}
    \par\bigskip
    \begin{subfigure}{\textwidth}
    	\includegraphics[width=\linewidth]{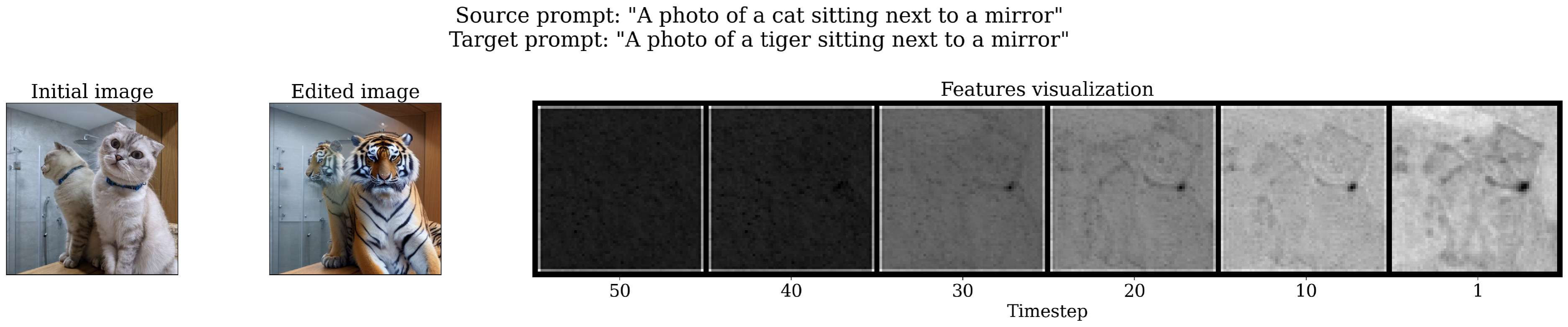}
    	\caption{Features $\Phi^*$ from the reference inversion trajectory $\varepsilon_{\theta}(z^*_t, t, y_{\mathrm{src}})$}
        \label{fig:feature_guider_visualization_non_styl_inv_inv}
    \end{subfigure}
    \vspace{-0.5cm}
    \caption{Visualization of \textit{features} from Feature Guider (Equation \ref{eq:features_other}) for non-stylisation editing example. We take mean over feature dimension of \textit{features}.}
    \label{fig:feature_guider_visualization_non_styl}
\end{figure*}

\begin{figure*}[t]
    \centering
    \begin{subfigure}{\textwidth}
    	\includegraphics[width=\linewidth]{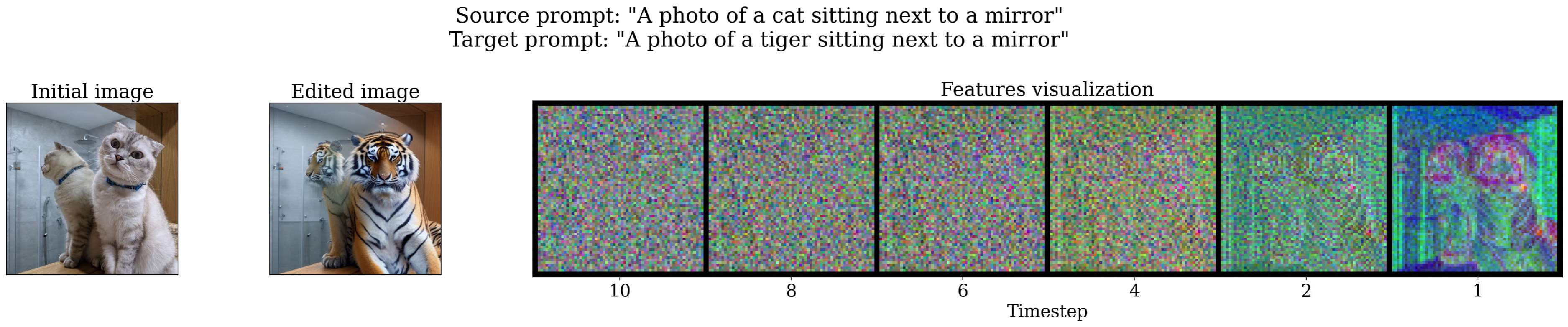}
    	\caption{Features $\bar{\Phi}$ from the current trajectory $\varepsilon_{\theta}(z_t, t, y_{\mathrm{src}})$}
        \label{fig:feature_guider_visualization_pca_non_styl_cur_inv}
    \end{subfigure}
    \par\bigskip
    \begin{subfigure}{\textwidth}
    	\includegraphics[width=\linewidth]{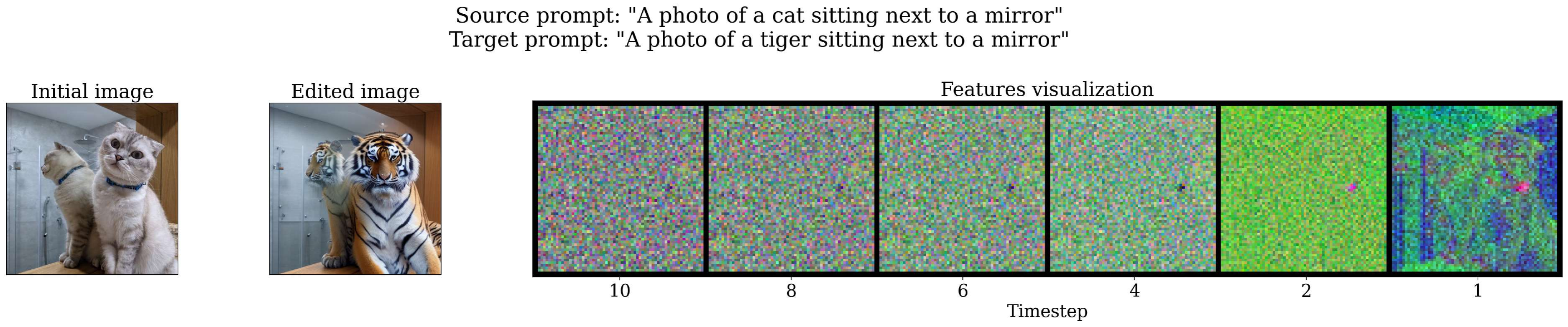}
    	\caption{Features $\Phi^*$ from the reference inversion trajectory $\varepsilon_{\theta}(z^*_t, t, y_{\mathrm{src}})$}
        \label{fig:feature_guider_visualization_pca_non_styl_inv_inv}
    \end{subfigure}
    \vspace{-0.5cm}
    \caption{Visualization of \textit{features} from Feature Guider (Equation \ref{eq:features_other}) for non-stylisation editing example. We show three leading principal components of \textit{features}.}
    \label{fig:feature_guider_visualization_pca_non_styl}
\end{figure*}

\begin{figure*}[t]
    \centering
    \begin{subfigure}{\textwidth}
    	\includegraphics[width=\linewidth]{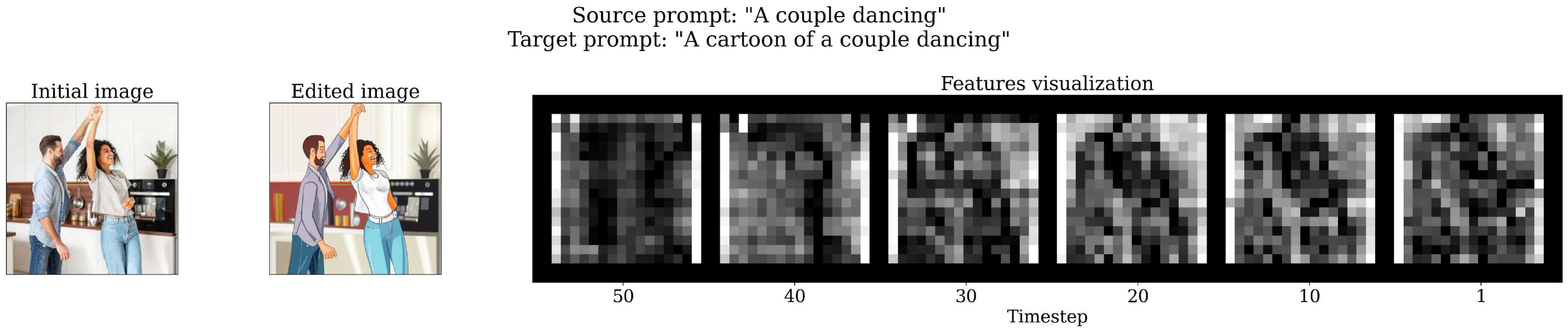}
    	\caption{Features $\bar{\Phi}$ from the current trajectory $\varepsilon_{\theta}(z_t, t, y_{\mathrm{trg}})$}
        \label{fig:feature_guider_visualization_styl_cur_trg}
    \end{subfigure}
    \par\bigskip
    \begin{subfigure}{\textwidth}
    	\includegraphics[width=\linewidth]{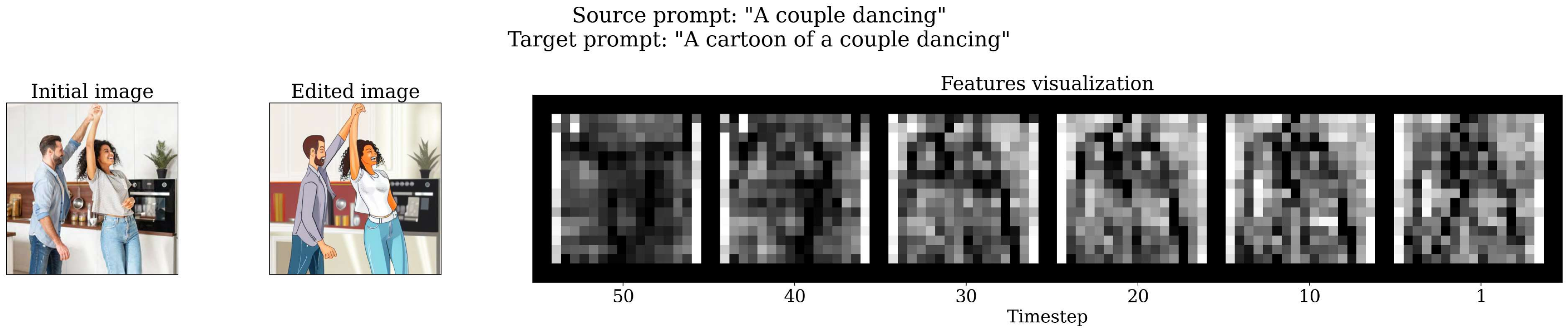}
    	\caption{Features $\Phi^*$ from the reference inversion trajectory $\varepsilon_{\theta}(z^*_t, t, y_{\mathrm{src}})$}
        \label{fig:feature_guider_visualization_styl_inv_inv}
    \end{subfigure}
    \vspace{-0.5cm}
    \caption{Visualization of \textit{features} from Feature Guider (Equation \ref{eq:features_stylization}) for stylisation editing example. We take mean over feature dimension of \textit{features}.}
    \label{fig:feature_guider_visualization_styl}
\end{figure*}

\begin{figure*}[t]
    \centering
    \begin{subfigure}{\textwidth}
    	\includegraphics[width=\linewidth]{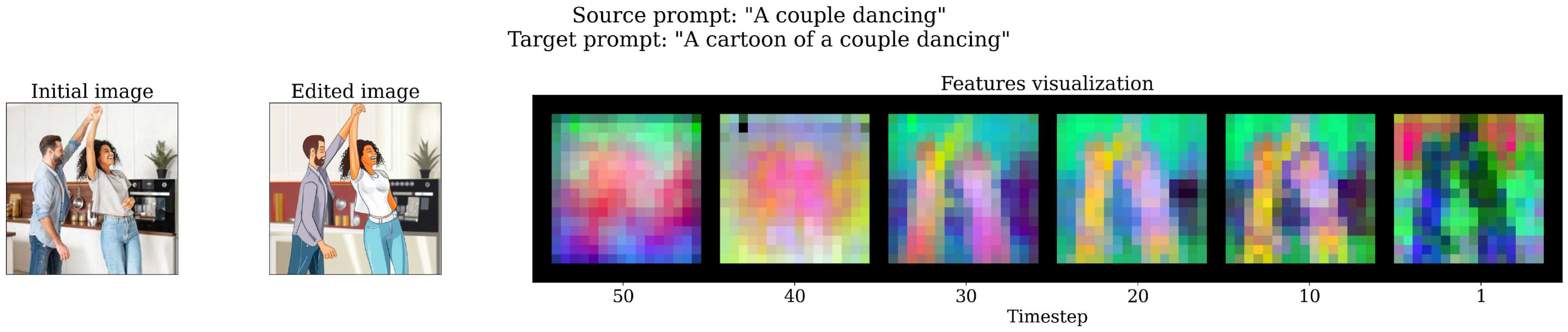}
    	\caption{Features $\bar{\Phi}$ from the current trajectory $\varepsilon_{\theta}(z_t, t, y_{\mathrm{trg}})$}
        \label{fig:feature_guider_visualization_pca_styl_cur_trg}
    \end{subfigure}
    \par\bigskip
    \begin{subfigure}{\textwidth}
    	\includegraphics[width=\linewidth]{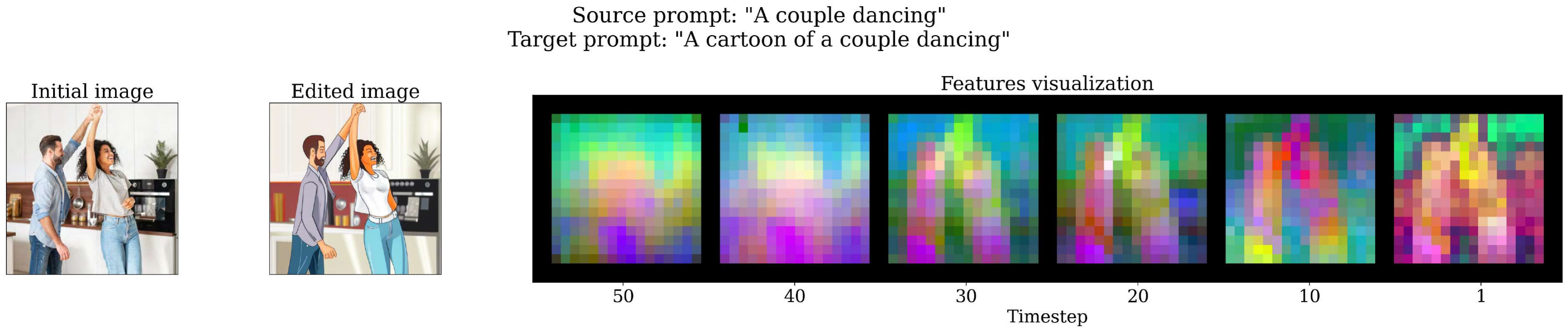}
    	\caption{Features $\Phi^*$ from the reference inversion trajectory $\varepsilon_{\theta}(z^*_t, t, y_{\mathrm{src}})$}
        \label{fig:feature_guider_visualization_pca_styl_inv_inv}
    \end{subfigure}
    \vspace{-0.5cm}
    \caption{Visualization of \textit{features} from Feature Guider (Equation \ref{eq:features_stylization}) for stylisation editing example. We show three leading principal components of \textit{features}.}
    \label{fig:feature_guider_visualization_pca_styl}
\end{figure*}

\noindent \textbf{Feature Guider.} This guider is defined differently for stylisation and non-stylisation editing tasks. It preserves the visual appearance of the whole picture. We provide examples of incorporating this guider into the editing pipeline in Fig. \ref{fig:feature_guidance_scale}. As in Fig. \ref{fig:self_attn_guidance_scale}, we show results of editing with CFG and Feature Guider, using different guidance scales:

\vspace{-0.3cm}
\begin{equation}
    \begin{array}{c}
        \epsilon_{\mathrm{final}} = \mathrm{CFG}(z_t, t, y_{\mathrm{trg}}, 7.5) + v \cdot \nabla g_{\mathrm{feat}}( \Phi^*, \bar{\Phi} ).
    \end{array}
\end{equation}

\noindent \textbf{Feature Guider for non-stylisation.} Formal definition is provided in Equation \ref{eq:features_other}. L1 norm in this guider is calculated similarly to Equation \ref{eq:l2_norm_def}:

\vspace{-0.3cm}
\begin{equation}\label{eq:l1_norm_def}
    \|X-Y\|_1 = \mathrm{abs}(X - Y).
\end{equation}

For non-stylisation editing, we consider the output of the last decoder layer in UNet as \textit{features}. We match \textit{features} from the current sampling trajectory, conditioning on $y_{\mathrm{src}}$: $\varepsilon_{\theta}(z_t, t, y_{\mathrm{src}})$, and the reference inversion trajectory $\varepsilon_{\theta}(z^*_t, t, y_{\mathrm{src}})$. We provide visualization of these features in Fig. \ref{fig:feature_guider_visualization_non_styl} and Fig. \ref{fig:feature_guider_visualization_pca_non_styl}. For visualization purposes, we take the mean over feature dimension of these \textit{features}, obtaining a matrix with spatial dimensions only, and visualize these matrices in Fig. \ref{fig:feature_guider_visualization_non_styl}. In Fig. \ref{fig:feature_guider_visualization_pca_non_styl}, we show three leading principal components of \textit{features}. This figure shows, that the same colors correspond to specific objects in the picture. However, the \textit{features} are too noisy during the first synthesis steps, so this pattern is hardly distinguishable.

\noindent \textbf{Feature Guider for stylisation.} Formal definition is provided in Equation \ref{eq:features_stylization}. This formula uses L2 norm defined in Equation \ref{eq:l2_norm_def}.

For stylisation editing, we consider the output of the second ResNet block in the second decoder layer in UNet as \textit{features}. In contrast to non-stylisation \textit{features}, for stylisation tasks, we define the current sampling trajectory as the one, conditioning on $y_{\mathbf{trg}}$: $\varepsilon_{\theta}(z_t, t, y_{\mathbf{trg}})$.

These features are visualized in Fig. \ref{fig:feature_guider_visualization_styl} and Fig. \ref{fig:feature_guider_visualization_pca_styl}.

\FloatBarrier

\section{Noise Rescaling}
\label{app:rescaling}

\begin{figure}[t]
    \includegraphics[width=\linewidth]{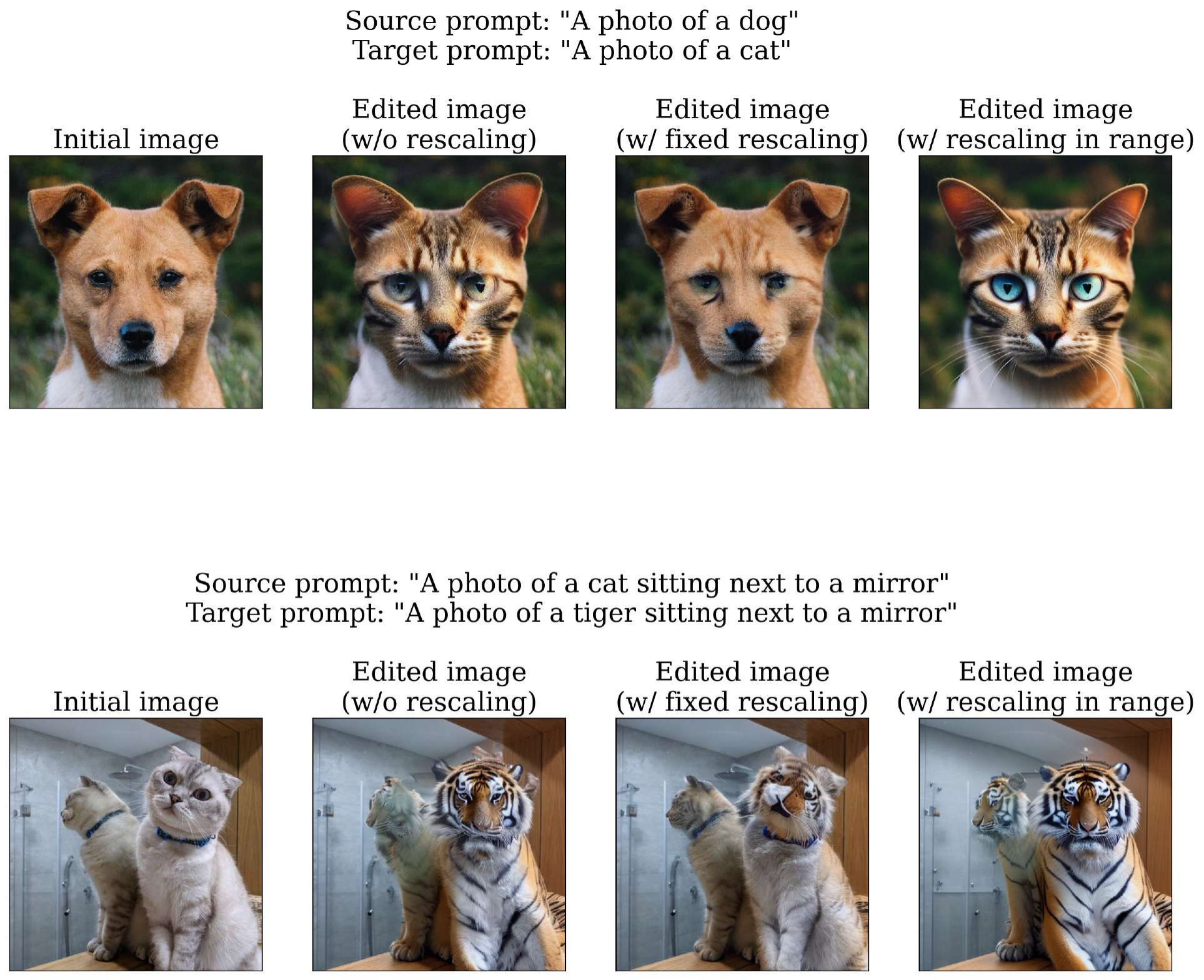}
    \caption{Illustration of applying different noise rescalings. Without rescaling there are some artifacts, with fixed rescaling the editing is not sufficiently visible, and with rescaling in a range the editing has a high quality.}
    \label{fig:rescaling_non_stylisation}
\end{figure}

\begin{figure}[t]
    \includegraphics[width=\linewidth]{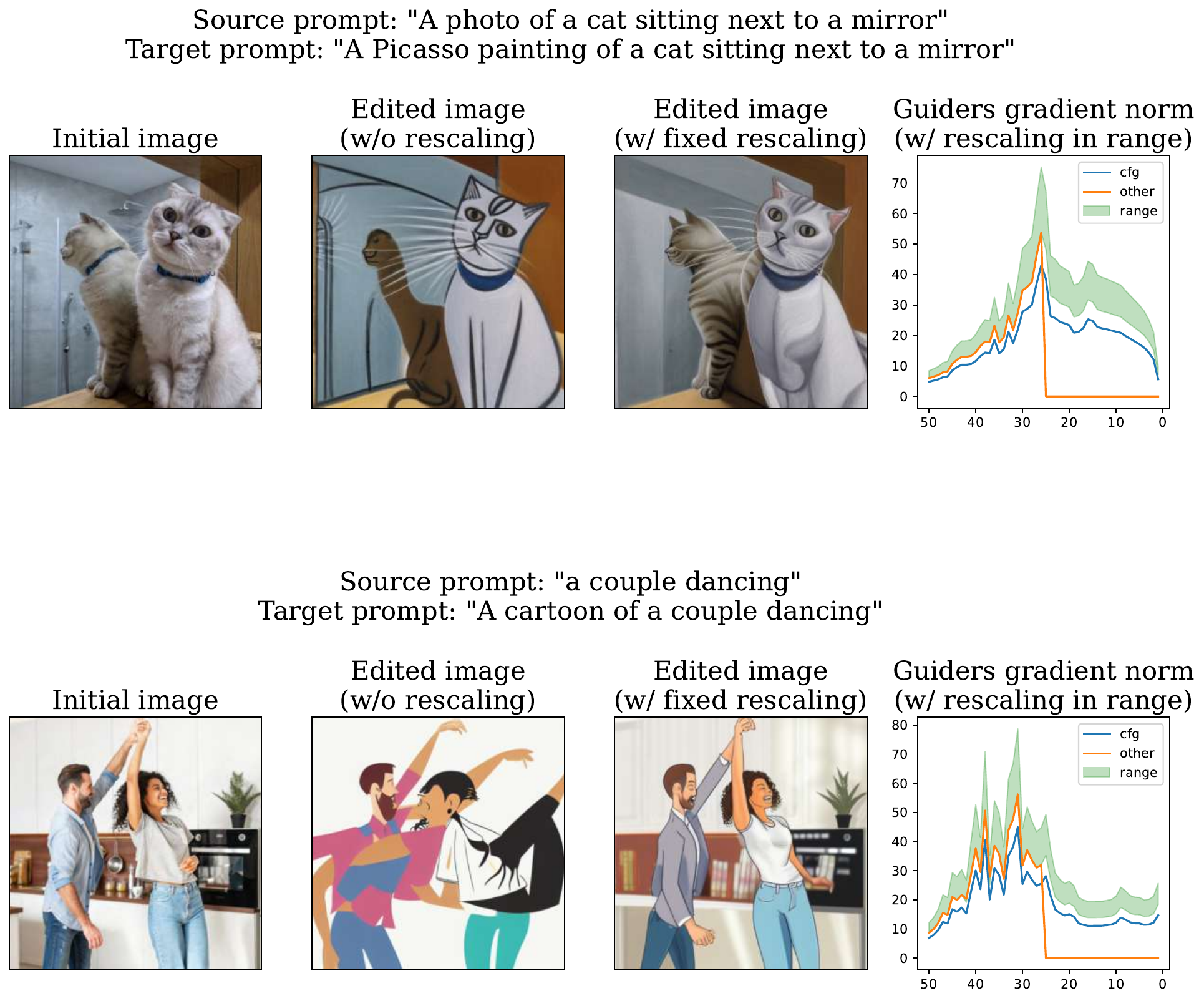}
    \caption{Illustration of applying different noise rescalings for stylisation. In the right plot, we show obtained gradient norms after rescaling in range and possible range for our guiders summary gradient norm.}
    \label{fig:rescaling_stylisation}
\end{figure}

In Section \ref{subsec:noise_rescaling} we define a noise rescaling technique that allows to control the balance between editing and preserving the content. We add Feature and Self-attention guiders to preserve the structure and the details of the image and use classifier-free guidance for editing. The greater the contribution of the guider in the sum of gradient norms, the greater the effect of that guider. To control the distribution of gradient norms, we add conditions on the gradient norm of guiders based on the gradient norm of CFG. 

At each iteration, we compute the relation $r_{\text{cur}}(t)$ between the gradient norms and then we evaluate the scaling factor $\gamma = r \cdot r_{\text{cur}}(t)$ as in Equation \ref{eq:scaling_factor_def}. We propose two approaches to choosing $r$.

\noindent \textbf{Fixed noise rescaling.} The first approach is to fix $r$ as some constant, for example take $r_{\text{fixed}} = 1.5$. This means that we rescale the gradient of the guiders in such a way that its norm is the scaled gradient norm of CFG at each iteration. For stylisation this is sufficient and it is shown in Fig. \ref{fig:rescaling_stylisation}, but for other types of editing, we propose using a more complicated noise rescaler.

\noindent \textbf{Noise rescaling in range.} Another way to rescale the noise is the fix $r$ in some range. The idea is to keep the relation between the sum of the gradient norms of the guiders and the CFG gradient norm, which is equal to $1/r_\text{cur}$ in the range $[r_\text{l}, r_\text{u}]$ and it means that $r_\text{cur}$ should be in the range $[1/r_\text{u}, 1/r_\text{l}]$. Hence, $r$ is

\begin{equation}\label{eq:r_evaluation}
    r = \left\{ \begin{array}{ll}
        r_{\mathrm{l}}, &  1/r_{\mathrm{cur}} \le r_{\mathrm{l}} \\
         1/ r_{\mathrm{cur}}, & r_{\mathrm{l}} < 1/ r_{\mathrm{cur}} < r_{\mathrm{u}}\\
         r_{\mathrm{u}}, & 1/r_{\mathrm{cur}} \ge r_{\mathrm{u}}
    \end{array} \right. .
\end{equation}

The scaling factor is then calculated similarly to Equation \ref{eq:scaling_in_range_appendix}. In Fig. \ref{fig:rescaling_non_stylisation} we show the difference between editing without rescaling, with fixed rescaling, and with rescaling in range. We show in Fig. \ref{fig:rescaling_stylisation} that we do not need to use rescaling in range for stylisation because the gradient norm of guiders is always less than the range and is clipped at the bottom.

It is worth mentioning, that by simply setting $r_{\mathrm{l}} = r_{\mathrm{u}}$ in Equation \ref{eq:r_evaluation} we can transform noise rescaling in range into fixed noise rescaling.

\section{Testing Datasets Description}
\label{app:dataset}

We compared our pipeline for both local and global editing with existing approaches such as NTI \cite{mokady2022null}, NPI \cite{miyake2023negativeprompt} and NPI Prox \cite{han2023improving} based on  P2P \cite{hertz2022prompt}, MasaCtrl \cite{cao_2023_masactrl}, ProxMasaCtrl \cite{han2023improving}, PnP \cite{Tumanyan_2023_CVPR}, and EDICT \cite{10204740}. We tested methods on several datasets. Section \ref{app:editing_types} describes our custom dataset consisting of four types of edits that we consider in our experiments. For each type, we manually assembled a dataset of 20 images. We also compared the methods on large datasets such as CoCo (Sec. \ref{app:coco_description}), AFHQ (Sec. \ref{app:afhq_description}) and Wild-FFHQ (Sec. \ref{app:ffhq_description}). 

- All results of these experiments are attached to the supplementary materials as separate pdf files.

\subsection{Custom Dataset Edit Types}
\label{app:editing_types}

\begin{figure*} 
    \includegraphics[width=\linewidth]{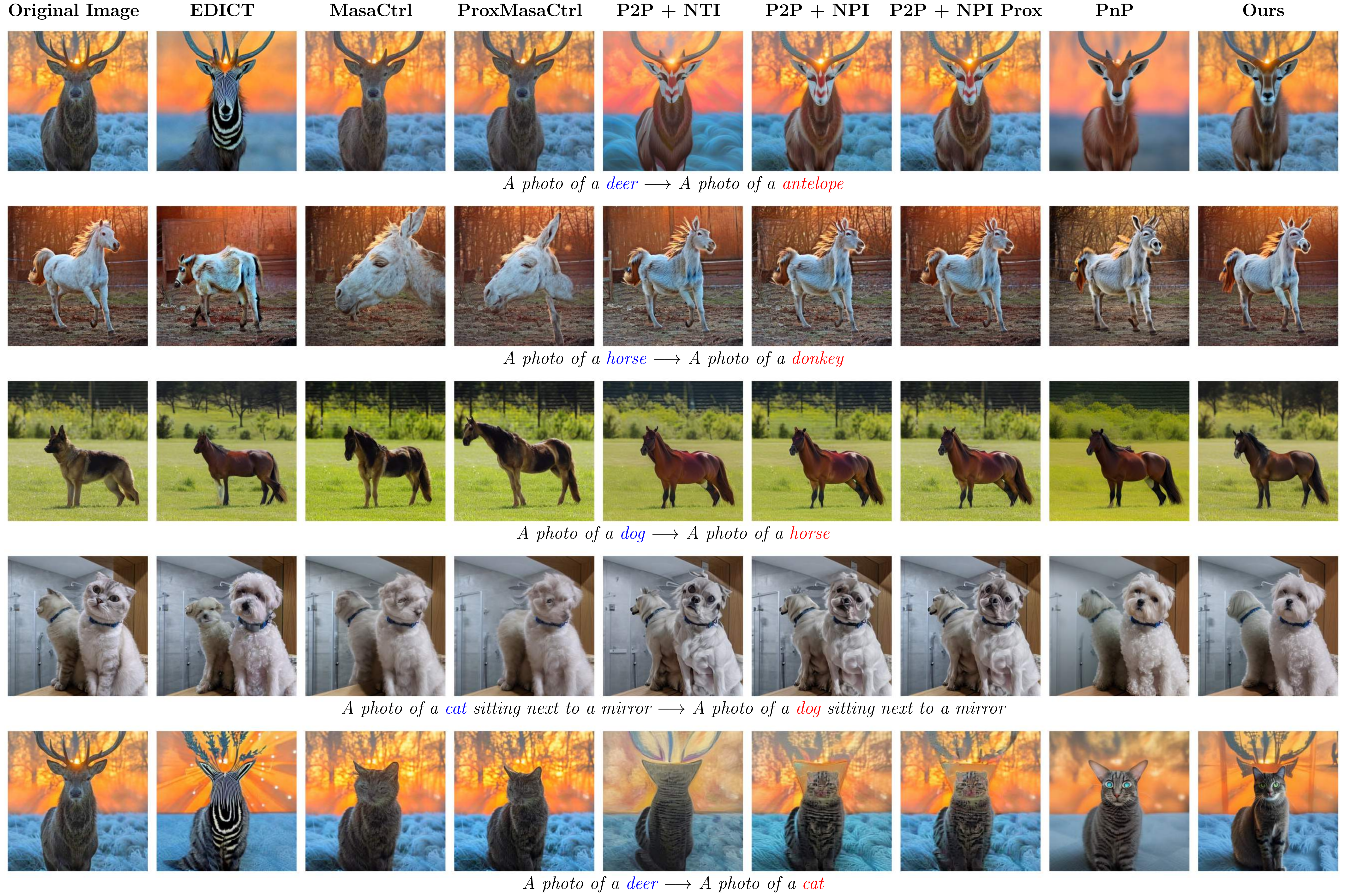}
    \caption{Visual comparison of our method vs baselines. The examples are from Custom Dataset for `animal-2-animal` type of editing.}
    \label{fig:visual_comparison_animal-2-animal}
\end{figure*}

\noindent \textbf{Animal to animal.} This is a local type of editing. The goal of this editing task is to change the depicted animal into a different one, preserving the background and the silhouette of the animal. Our test set contains pictures of various animals (cat, dog, tiger, horse, deer), some of which have very specific features (for example, a deer has antlers). We consider changing an animal into the one, having a naturally similar shape (for example, changing a tiger into a panther), making complicated swaps (for example, changing a deer into a cat), and changing the color of the animal with preserving the majority of the features (changing a white horse into a black one). Visual comparison for some of these examples is reported in Fig. \ref{fig:visual_comparison_animal-2-animal}.

\begin{figure*} 
    \includegraphics[width=\linewidth]{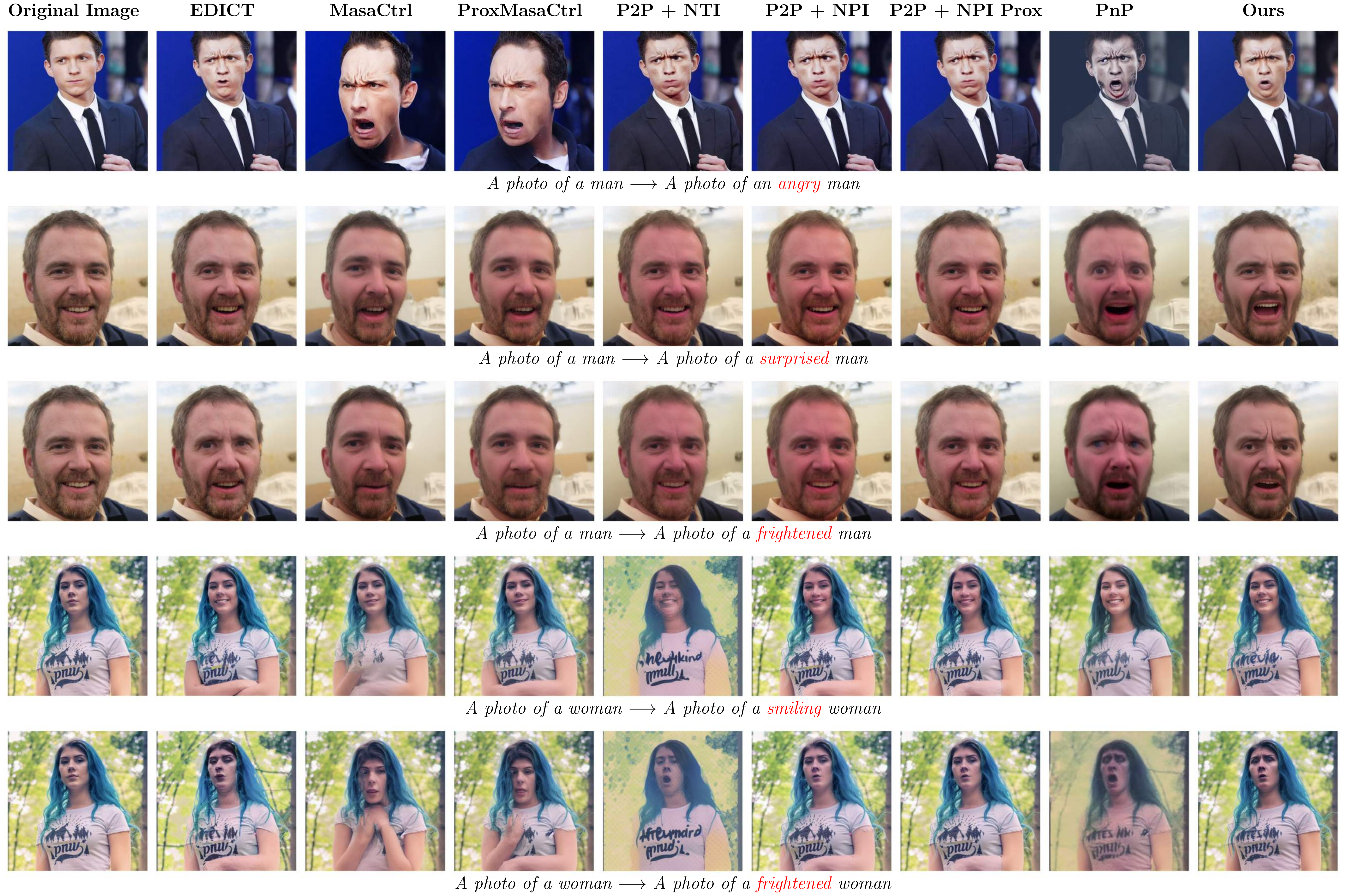}
    \caption{Visual comparison of our method vs baselines. The examples are from Custom Dataset for `face-in-the-wild` type of editing.}
    \label{fig:visual_comparison_face-in-the-wild}
\end{figure*}

\noindent \textbf{Face in the wild.} This is a local type of editing. In our setting, this task aims at changing the emotion of the person in the picture and preserving the background and the appearance of the person (meaning that the person in the edited picture should still resemble the person in the initial image). We conduct experiments on photographs of different people, taken from different perspectives. Our initial images depict people with different emotions as well (neutral and smiling). However, we still obtain good results in this task by not specifying the emotion in the source prompt at all. Every source prompt has a form of ``A photo of a \{man/woman\}". The target prompt can be easily obtained from the source prompt by adding the desired emotion (for example, happy, smiling, crying) prior to the word ``\{man/woman\}". Visual comparison for some of these examples is reported in Fig. \ref{fig:visual_comparison_face-in-the-wild}.

\begin{figure*} 
    \includegraphics[width=\linewidth]{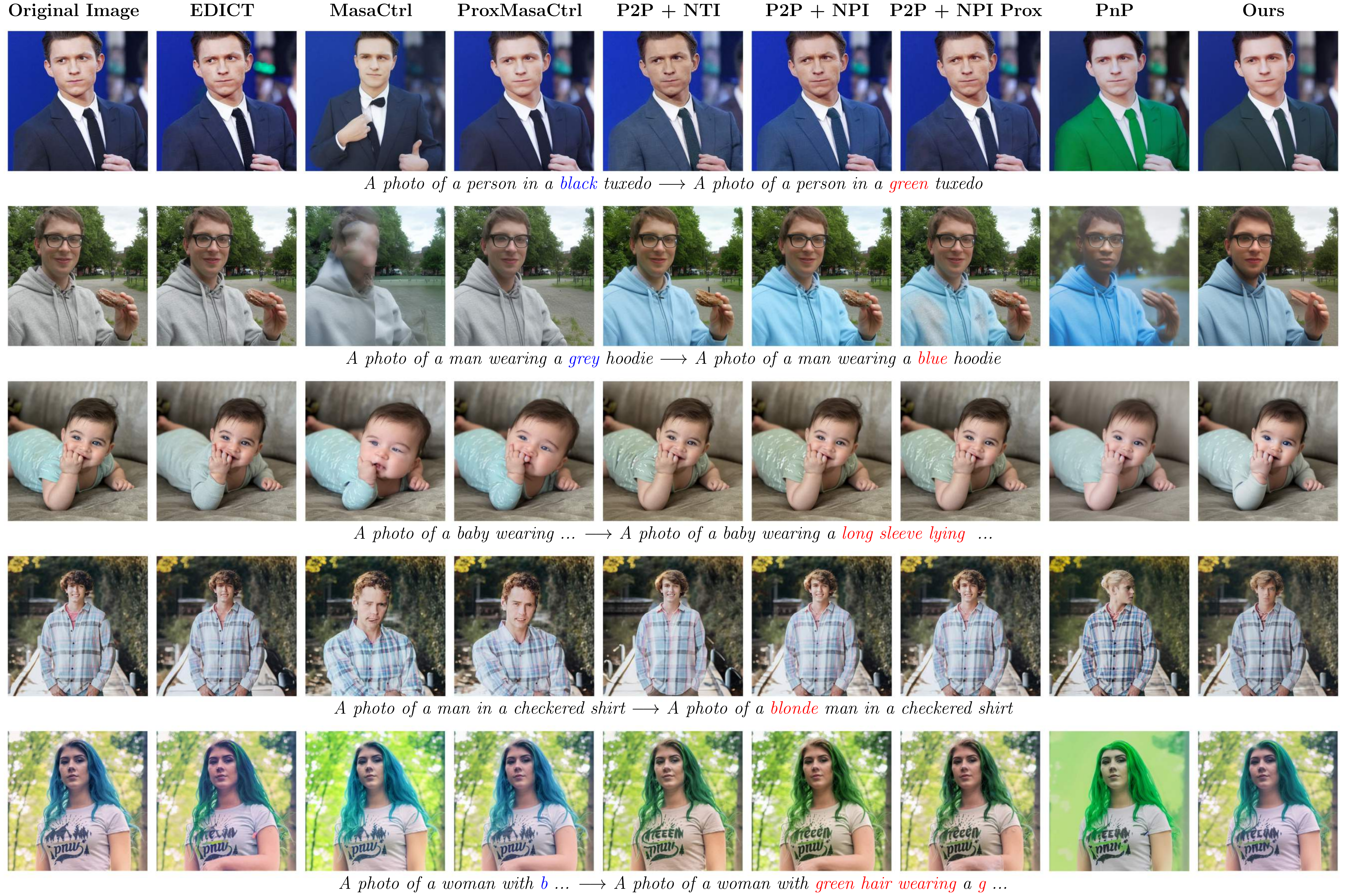}
    \caption{Visual comparison of our method vs baselines. The examples are from Custom Dataset for `person-in-the-wild` type of editing.}
    \label{fig:visual_comparison_person-in-the-wild}
\end{figure*}

\begin{figure*} 
    \includegraphics[width=\linewidth]{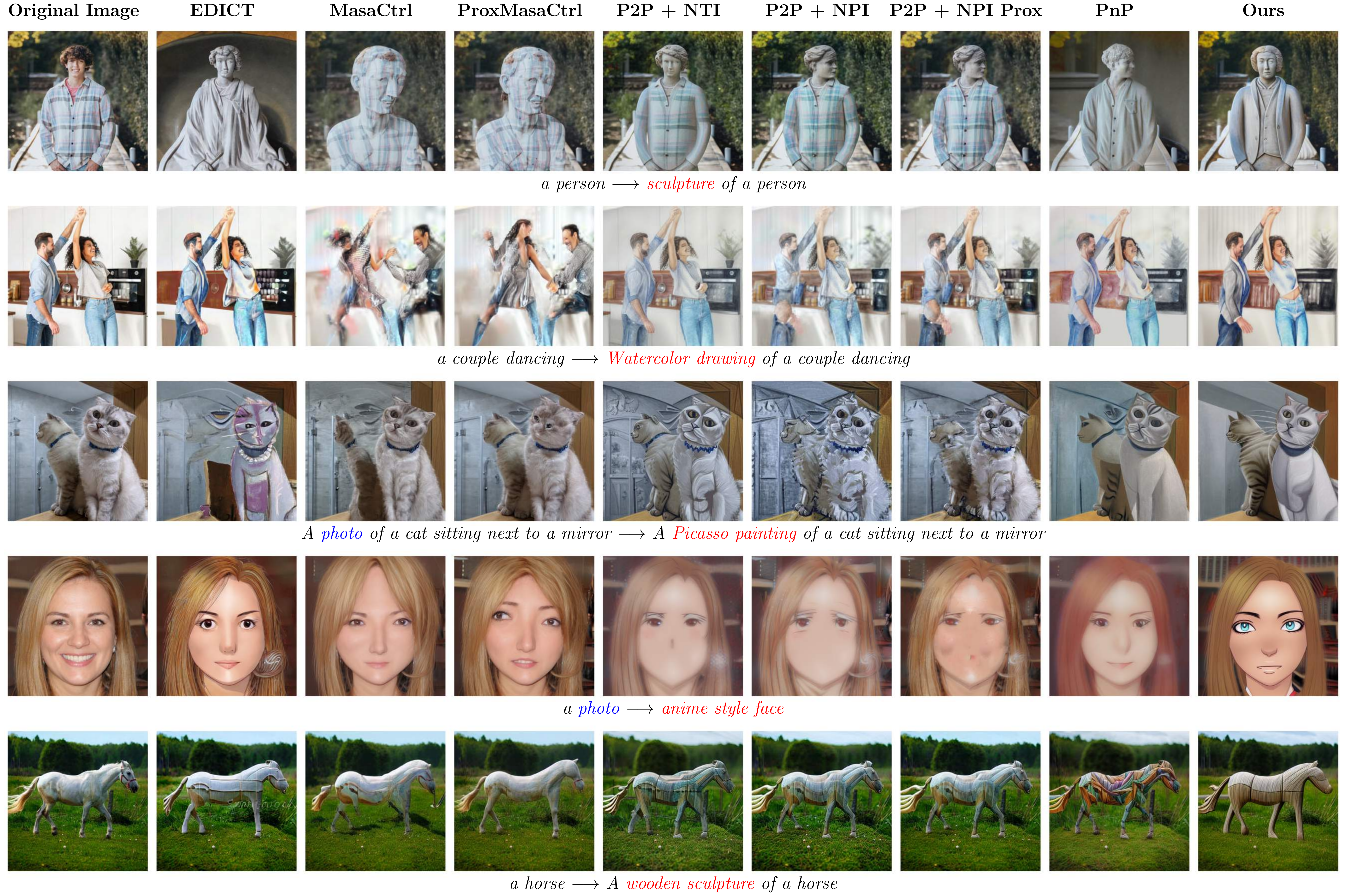}
    \caption{Visual comparison of our method vs baselines. The examples are from Custom Dataset for `stylisation` type of editing..}
    \label{fig:visual_comparison_stylisation}
\end{figure*}

\noindent \textbf{Stylisation.} This is a global type of editing. We tested the methods on different drawing styles (sketch, oil, watercolor), artist stylisations (Van Gogh, Picasso, Fernando Botero), changing the texture of objects (wooden statues and sculptures), changing face styles (anime, pixar, pop art). Visual comparison for some of these examples is reported in Fig. \ref{fig:visual_comparison_stylisation}.

\subsection{Stylisation - CoCo dataset}
\label{app:coco_description}
The MS COCO (Microsoft Common Objects in Context) dataset \cite{lin2015microsoft} is a large-scale dataset with 164K high-resolution images and annotations to them. We use a subset of 500 images from this dataset for the stylisation type of editing. 
Each testing object is constructed with a randomly picked image. The source prompt is the corresponding annotation for the picked image. The target prompt is constructed by concatenating the random style from the list of styles with the source prompt. 

We randomly choose a style for each image from the list: 'Anime Style', 'A pop art style', 'A pixar style', 'A Van Gogh painting of', 'A Fernando Botero painting of', 'A Ukiyo-e painting of', 'A Picasso painting of', 'A charocal painting of', 'An oil painting of', 'A sketch of', 'A cubism painting of', 'An impressionist painting of', 'A watercolor drawing of', 'A minecraft painting of', 'Banksy art of', 'da Vinci sketch of', 'A mosaic depicting of', 'A gothic painting of', 'A geometric abstract painting of', 'A white wool of', 'A futurism painting of', 'A Pixel art style', 'Comic book style', 'Cyberpunk style', 'Flat style'.

\begin{figure*} 
    \includegraphics[width=\linewidth]{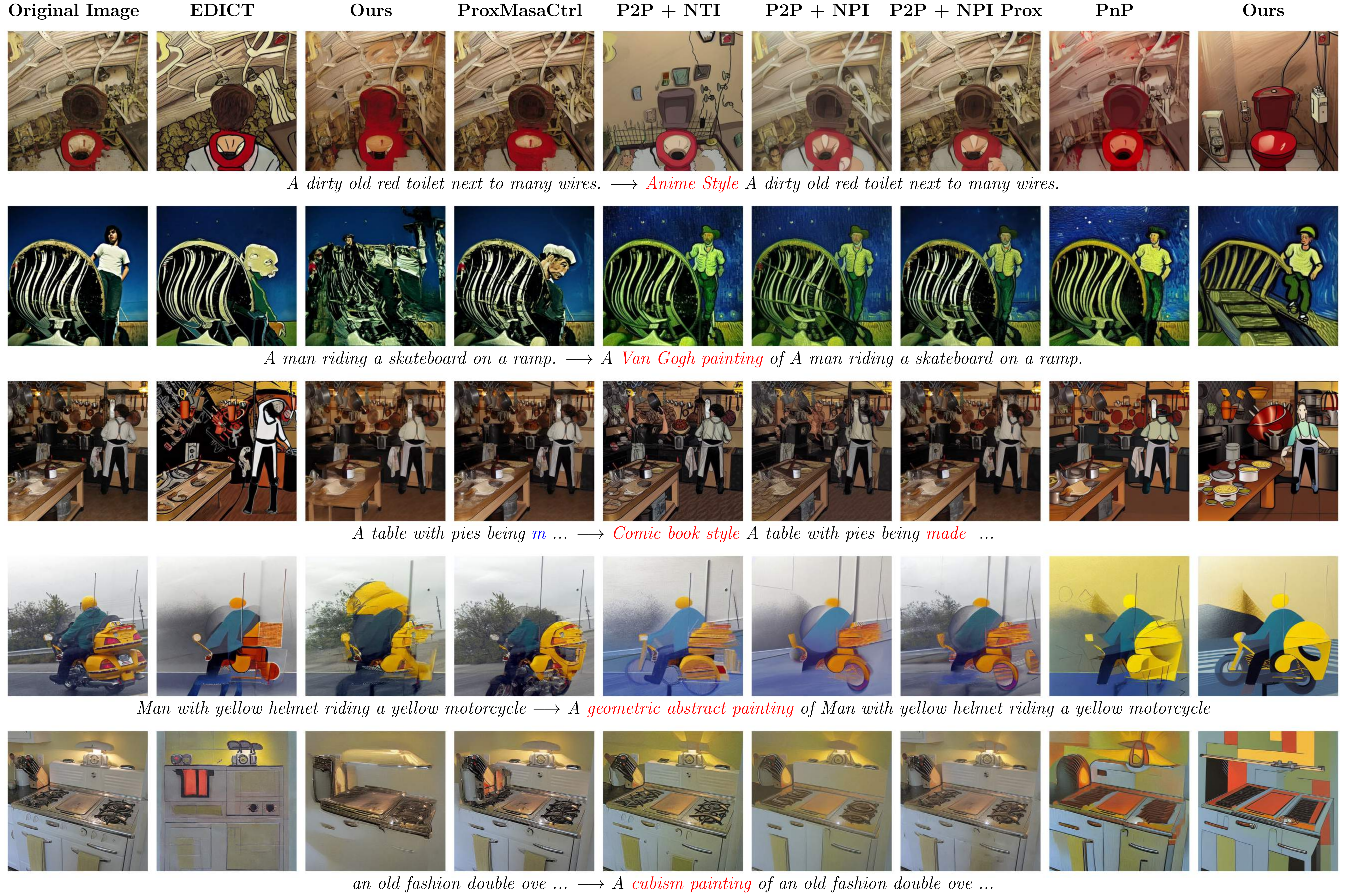}
    \caption{Visual comparison of our method vs baselines. The examples are from MS COCO dataset for `stylisation` type of editing.}
    \label{fig:visual_comparison_coco_stylisation}
\end{figure*}

We show visual comparison in the Fig. \ref{fig:visual_comparison_coco_stylisation}. 

\subsection{Dog2Cat - AFHQ}
\label{app:afhq_description}
Animal FacesHQ (AFHQ) \cite{choi2020starganv2} is a dataset of 15k high-quality images at 512x512 resolution of animal faces. It contains three types of animals: cats, dogs, and wild animals. We use a subset of 500 images of dogs from this dataset for the dog-to-cat type of editing. 
Each testing object is constructed with a randomly picked image. We use 'a dog' as a source prompt and 'a cat' as a target prompt for the transformation of dogs into cats. We report FID in Table \ref{tab:quant_comparison}. To evaluate one we use a subset of 5k images of cats to compare the generated distribution and the original one.

\begin{figure*} 
    \includegraphics[width=\linewidth]{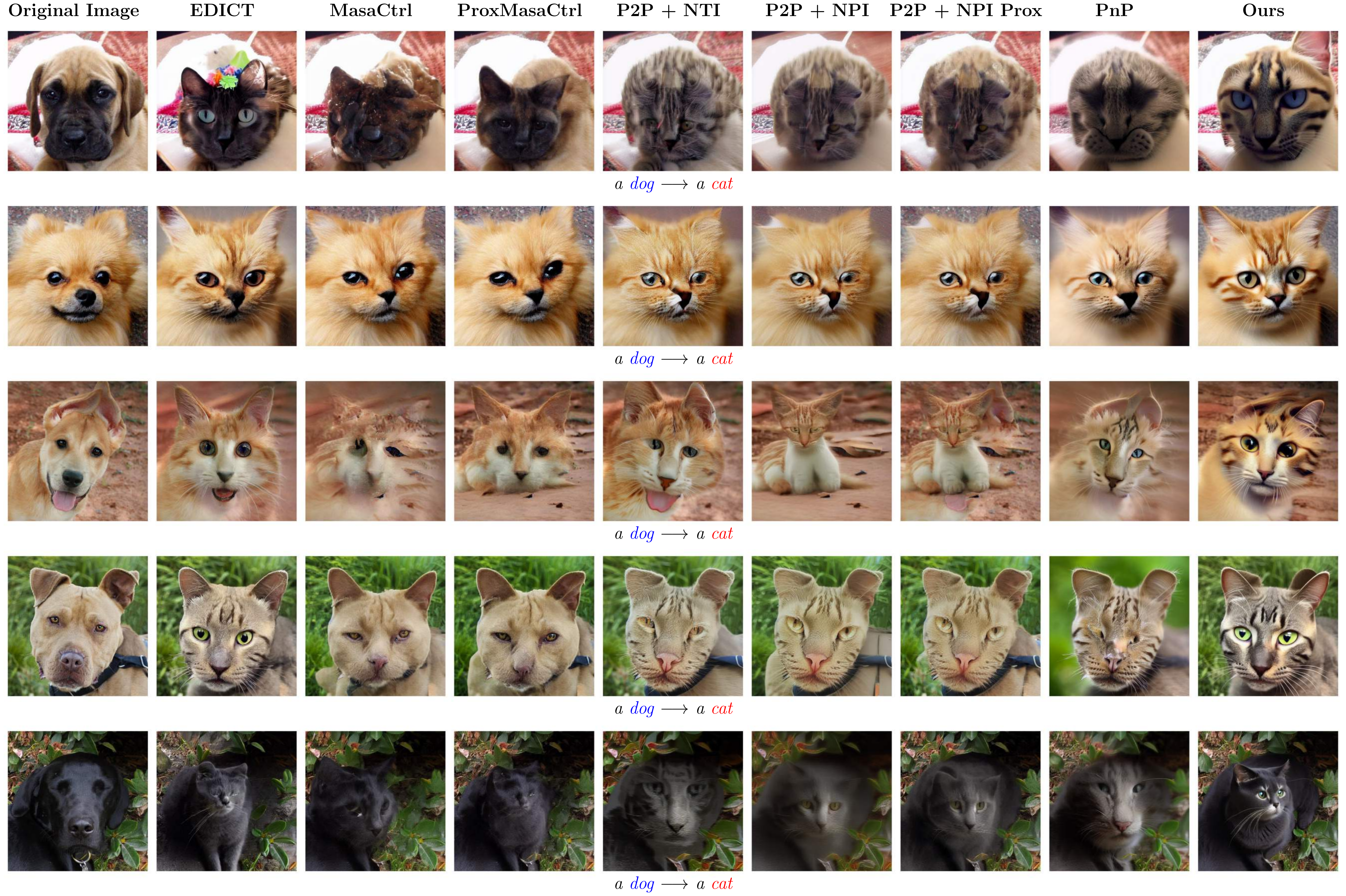}
    \caption{Visual comparison of our method vs baselines. The examples are from AFHQ dataset for `dog-to-cat` type of editing.}
    \label{fig:visual_comparison_dog_to_cat}
\end{figure*}

We show visual comparison in the Fig. \ref{fig:visual_comparison_dog_to_cat}. 

\subsection{Change of emotions - Wild FFHQ}
\label{app:ffhq_description}

We use a subset of 500 images from FFHQ dataset \cite{Karras2019ASG}. Each testing object is constructed with a wild face example from FFHQ. The source prompt is "A photo of a person". To construct the target prompt we used the following procedure: we randomly picked one of the emotion descriptions and inserted it into the prompt.

The set of emotion descriptions: 'frightened', 'laughing', 'shy', 'surprised', 'smiling', 'crying', 'angry', 'sad', 'happy', 'emotionless', 'disgusted', 'painful', 'thoughtful', 'worried', 'curious'

\begin{figure*} 
    \includegraphics[width=\linewidth]{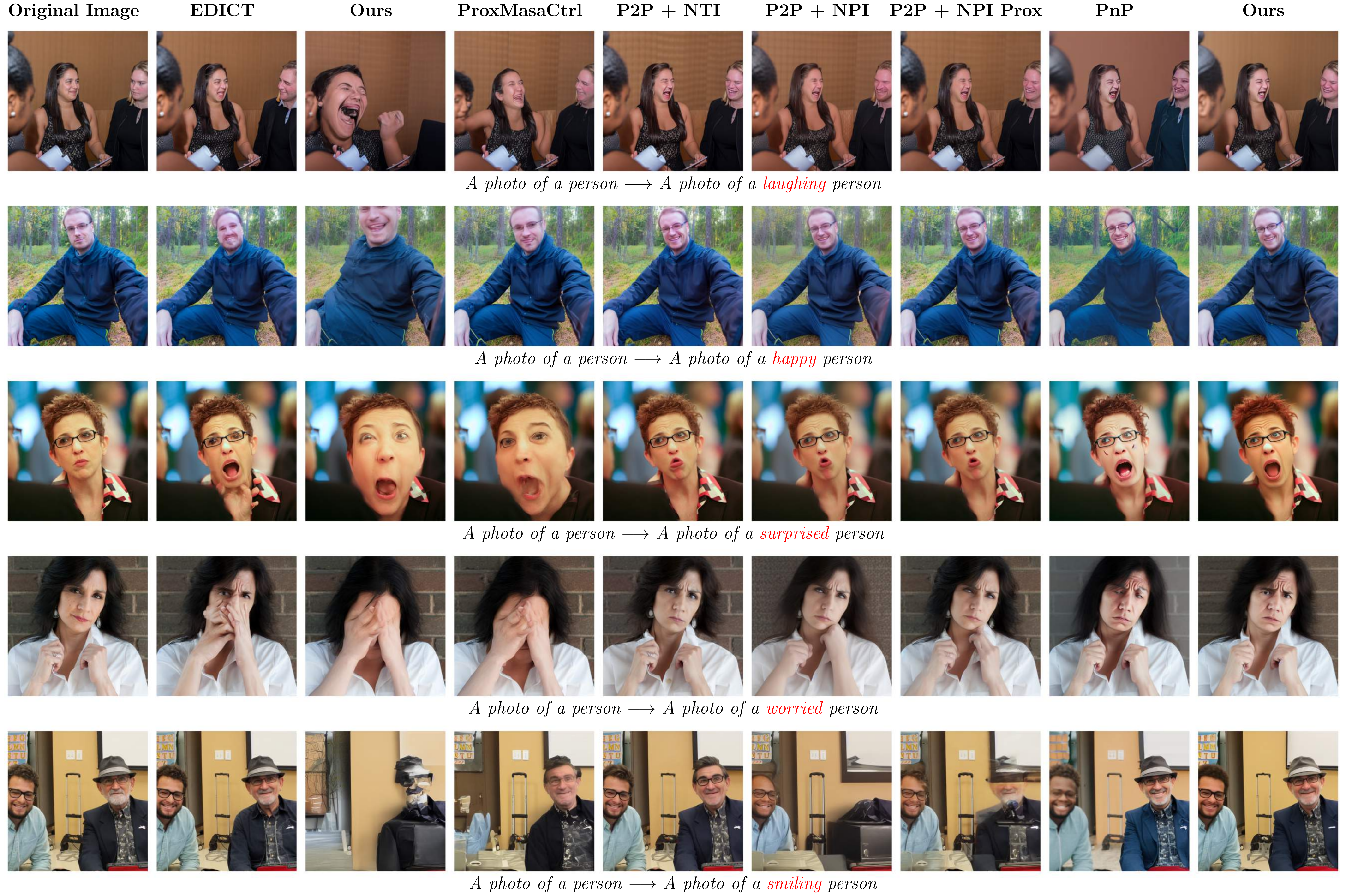}
    \caption{Visual comparison of our method vs baselines. The examples are from Wild FFHQ dataset for `face-in-the-wild` type of editing.}
    \label{fig:visual_comparison_ffhq_wild}
\end{figure*}

For that dataset, we show visual comparison in Fig. \ref{fig:visual_comparison_ffhq_wild}.

\FloatBarrier

\section{Method Setup for Experiments}
\label{app:setup}

\newcommand{\pnpcode}{\href{https://github.com/MichalGeyer/plug-and-play}{PnP github-repo}}

\newcommand{\masactrlcode}{\href{https://github.com/TencentARC/MasaCtrl}{MasaCTRL github-repo}}

\newcommand{\proxgithub}{\href{https://github.com/phymhan/prompt-to-prompt}{Prox github-repo}}

\newcommand{\ninvptopcode}{\href{https://github.com/google/prompt-to-prompt}{P2P github-repo}}

\newcommand{\edictcode}{\href{https://github.com/salesforce/EDICT}{Edict github-repo}}

\begin{table*}[h!]
    \small
    \centering
    \begin{tabular}{l c c}
        \toprule
        Method Name & Used Implementation & Special Parameters \\ \hline
        Plug-and-Play & \pnpcode & \makecell{self\_attn\_block\_indices = [4, 5, 6, 7, 8, 9, 10, 11] \\
        out\_layers\_output\_block\_indices = [4] \\
        feature\_injection\_threshold = 40 \\
        scale = 10.
        } \\
        \hline
        Edict & \edictcode & \makecell{
        mix\_weight = 0.93 \\
        init\_image\_strength = 0.8 \\
        guidance\_scale = 3 \\
        } \\
        \hline
        MasaCTRL & \masactrlcode & \makecell{
        STEP = 4 \\
        LAYPER = 10 \\
        } \\ 
        \hline
        ProxMasaCTRL & \proxgithub & \makecell{
        npi = True \\
        npi\_interp = 1 \\
        prox = 'l0' \\
        quantile = 0.6 \\
        masa\_step = 4 \\
        masa\_layer = 10 \\
        } \\
        \hline
        NPI + p2p & \ninvptopcode & \makecell{
        cross\_replace\_steps = 0.8 \\
        self\_replace\_steps = 0.6 \\
        blend\_word = None
        } \\
        \hline
        NPI prox + p2p & \proxgithub & \makecell{
        cross\_replace\_steps = 0.8 \\
        self\_replace\_steps = 0.6 \\
        blend\_word = None \\
        prox = 'l0' \\
        quantile = 0.6 \\
        } \\
        \hline
        Null-Text + p2p & \ninvptopcode & \makecell{
        cross\_replace\_steps = 0.8 \\
        self\_replace\_steps = 0.6 \\
        blend\_word = None \\
        opt\_early\_stop\_epsilon (NullInv) = 1e-5 \\
        opt\_num\_inner\_steps (NullInv) = 10 
        } \\
        \bottomrule
    \end{tabular}
    \caption{Special arguments that were used while inference of baseline methods when comparing. Also, source repositories are provided.}
    \label{table:baselines_setup}
\end{table*}

\FloatBarrier

\section{Method Setup for Experiments}
\label{app:setup}

All experiments on methods comparison were conducted using authors' official repositories except NPI \cite{miyake2023negativeprompt} - for this method modified p2p code has been used. For each editing type of test example, default parameters were used.
All the corresponding code implementation and special parameters of the method's inference can be found in Table \ref{table:baselines_setup}.

The setup for our method can be found in Sec. \ref{app:pipeline}. 

\FloatBarrier

\section{Quantitative Evaluation}
\label{app:quant}

For quantitative comparison, paired-metric comparison is provided. That comparison is previously used in EDICT method \cite{10204740}. 

We quantitatively benchmark all methods on 4 test datasets: Custom Dataset, CoCo, Wild-FFHQ, AFHQ. An exact explanation of the way we were using each dataset can be found in Sec. \ref{app:dataset}.

LPIPS \cite{zhang2018unreasonable} metric is used to analyze original image preservation properties, CLIP score \cite{radford2021learning} is used to analyze semantic similarity to desired editing description. The motivation for using LPIPS metric is in the perception of one to visual changes. In common editing problems (except stylisation) there is a requirement to save non-editing areas unchanged so low values of LPIPS indicate that the method preserves the original structure of the image better. The motivation for using CLIP score is in its perception of the similarity of editing results to target editing description. As it follows by the motivation of metrics, the effective method will have lower LPIPS and higher CLIP score. Results for the custom dataset are shown in Table \ref{tab:quant_comparison}. For CoCo results are shown in Table \ref{tab:coco_quant}. For FFHQ results are shown in Table \ref{tab:ffhq_quant}.

Our method greatly outperforms \cite{cao_2023_masactrl, han2023improving} in both metrics, and confirmation of it is in visual comparisons. Compared to p2p method \cite{hertz2022prompt, mokady2022null, miyake2023negativeprompt} our method is also superior in both metrics but with less gap, such behavior is aligned with visual results, as p2p works better than MasaCTRL. Compared to Edict \cite{10204740} our method has comparable LPIPS value but vastly superior CLIP score value. In contrast, compared to PnP \cite{Tumanyan_2023_CVPR} our method has a comparable CLIP score value but a significantly better LPIPS value. To prove the superiority of our method over others we propose to average rank for each metric and analyze its average rank. We have introduced a metric called AverageRank to show comparison based on a single value, which is a final indicator of the quality of the method. That average ranking shows the superiority of our method over other baselines. Results are shown in Table \ref{table:average_rank}. 

For the CoCo-`stylisation` problem there our method shows poor LPIPS because our result shows better stylisation and loses identical spatial edges that make LPIPS less. For the FFHQ problem, our method shows an optimal LPIPS score while having quite high CLIP score.

Also, we are examining our method on classic dog-to-cat problem. To estimate the quality of our method and compare it with other baselines we used data described in Sec. \ref{app:afhq_description}. The main results on FID score are reported in Table \ref{tab:quant_comparison}. Our method outperforms all other baseline methods.

\begin{table*}[t]
    \small
    \centering
    \begin{tabular}{c | c | c | c | c}
    Method Name & LPIPS Rank $\downarrow$ & CLIPscore Rank $\downarrow$ & FID Rank $\downarrow$ & Average Rank $\downarrow$ \\ 
    \midrule
Guide-and-Rescale (ours) & 3 (0.228) & 2 (0.243) & 1 (39.07) & 2.0 \\
P2P \cite{hertz2022prompt} + NPI \cite{miyake2023negativeprompt} Prox \cite{han2023improving} & 1 (0.170) & 4 (0.233) & 4 (43.16) & 3.0 \\
PnP \cite{Tumanyan_2023_CVPR} & 8 (0.366) & 1 (0.256) & 2 (39.55) & 3.7 \\
P2P \cite{hertz2022prompt} + NPI \cite{miyake2023negativeprompt} & 4 (0.251) & 3 (0.234) & 5 (44.05) & 4.0 \\
EDICT \cite{10204740} & 2 (0.221) & 6 (0.229) & 6 (47.13) & 4.7 \\
P2P \cite{hertz2022prompt} + NTI \cite{mokady2022null} & 6 (0.279) & 5 (0.233) & 3 (42.46) & 4.7 \\
ProxMasaCtrl \cite{han2023improving} & 5 (0.267) & 8 (0.215) & 7 (94.53) & 6.7 \\
MasaCtrl \cite{cao_2023_masactrl} & 7 (0.306) & 7 (0.223) & 8 (100.62) & 7.3 \\
    \hline
    \end{tabular}
    \caption{Comparison overall methods based on introduced AverageRank metric. Each rank shows the method's position in a sorted metrics array. In brackets, there are the exact values of the corresponding metric.}
    \label{table:average_rank}
\end{table*}

\begin{table}[h!]
    \vspace{-0.2cm}
    \centering
    \caption{Comparison with baselines on CoCo for stylisation editing task.}
    \small
    \begin{tabular}{lcc}
    \toprule
        \textbf{\quad \ \ \ Method} & LPIPS $\downarrow$ & CLIP $\uparrow$ \\
        \midrule
MasaCtrl \cite{cao_2023_masactrl} &  0.300 &  0.240 \\
ProxMasaCtrl \cite{han2023improving} &  0.227 &  0.231 \\
P2P \cite{hertz2022prompt} + NTI \cite{mokady2022null} &  0.338 &  0.272 \\
P2P \cite{hertz2022prompt} + NPI \cite{miyake2023negativeprompt} &  0.394 &  0.265 \\
P2P \cite{hertz2022prompt} + NPI \cite{miyake2023negativeprompt} Prox \cite{han2023improving} &  \textbf{0.208} &  0.263 \\
EDICT \cite{10204740} &  0.273 &  0.253 \\
PnP \cite{Tumanyan_2023_CVPR} &  0.378 &  0.286 \\
\midrule
Guide-and-Rescale (ours) &  0.449 &  \textbf{0.290} \\
        \bottomrule
    \end{tabular}
    \label{tab:coco_quant}
    \vspace{-0.2cm}
\end{table}

\begin{table}[h!]
    \centering
    \caption{Comparison with baselines on FFHQ for `change of emotions` editing task.}
    \small
    \begin{tabular}{lcc}
    \toprule
        \textbf{\quad \ \ \ Method} & LPIPS $\downarrow$ & CLIP $\uparrow$ \\
        \midrule
MasaCtrl \cite{cao_2023_masactrl} &  0.346 &  0.190 \\
ProxMasaCtrl \cite{han2023improving} &  0.281 &  0.184 \\
P2P \cite{hertz2022prompt} + NTI \cite{mokady2022null} &  \textbf{0.120} &  0.171 \\
P2P \cite{hertz2022prompt} + NPI \cite{miyake2023negativeprompt} &  0.249 &  0.175 \\
P2P \cite{hertz2022prompt} + NPI \cite{miyake2023negativeprompt} Prox \cite{han2023improving} &  0.134 &  0.171 \\
EDICT \cite{10204740} &  0.158 &  0.174 \\
PnP \cite{Tumanyan_2023_CVPR} &  0.251 &  \textbf{0.190} \\
\midrule
Guide-and-Rescale (ours) &  0.156 &  0.183 \\
        \bottomrule
    \end{tabular}
    \label{tab:ffhq_quant}
\end{table}

\FloatBarrier

\section{User Study}
\label{app:user_study}

We have provided user studies experiment with the dataset described in Sec. \ref{app:editing_types}. For each method mentioned in Table \ref{tab:user_study} you can find inference hyperparameters described in Sec. \ref{app:setup}.

\subsection{Experimental setup}
\label{app:user_studies_setup}

For each user, there were shown the original image, the description of the desired editing, the result of our method, and the result of the baseline method. 

Users had to answer two questions. The first question (Q1 - Edit property) is "Which image matches the desired editing description best?". The second question (Q2 - Preservation property) is "Which image preserves the overall structure of the `Original Image` best?". The user task consisted of three comparisons. 

For each object from the dataset, there were seven pairs of comparison: Ours vs each of the baseline. To decrease user's variance and improve the stability of user studies we set overlap for each unique testing object to 3. This means that every unique comparison was shown to 3 unique users. Also to decrease bias of user study results number of attempts for tasks for each user was limited to 6. 

\subsection{Metrics per editing type of Custom Dataset}

Results for all types of editing are shown in Table \ref{tab:user_study}. Results over each editing type are shown in Table [\ref{tab:user_study_face}, \ref{tab:user_study_a2a}, \ref{tab:user_study_style}, \ref{tab:user_study_person}].

\begin{table}[h!]
    \centering
    \caption{
    User studies that show the percentage of users that preferred our method. The current table shows a comparison on the subset of results that belong to \textbf{face-in-the-wild} editing type.}
    \label{tab:user_study_face}
    \small
    \begin{tabular}{lccc}
    \toprule
        \textbf{\quad \ \ \ Method}         & Preservation  & Editing \\
        \midrule
P2P \cite{hertz2022prompt} + NTI \cite{mokady2022null} & 40 \% & 73 \% \\
P2P \cite{hertz2022prompt} + NPI \cite{miyake2023negativeprompt} Prox \cite{han2023improving} & 32 \% & 47 \% \\
P2P \cite{hertz2022prompt} + NPI \cite{miyake2023negativeprompt} & 67 \% & 83 \% \\
PnP \cite{Tumanyan_2023_CVPR} & 72 \% & 89 \% \\
EDICT \cite{10204740} & 68 \% & 68 \% \\
ProxMasaCtrl \cite{han2023improving} & 67 \% & 83 \% \\
MasaCtrl \cite{cao_2023_masactrl} & 75 \% & 25 \% \\
        \bottomrule
    \end{tabular}
\end{table}

\begin{table}[h!]
    \centering
    \caption{User studies that show the percentage of users that preferred our method. The current table shows the comparison on the subset of results that belong to \textbf{animal-2-animal} editing type.}
    \small
    \begin{tabular}{lccc}
    \toprule
        \textbf{\quad \ \ \ Method}         & Preservation  & Editing \\
\midrule
P2P \cite{hertz2022prompt} + NTI \cite{mokady2022null} & 67 \% & 59 \% \\
P2P \cite{hertz2022prompt} + NPI \cite{miyake2023negativeprompt} Prox \cite{han2023improving} & 92 \% & 67 \% \\
P2P \cite{hertz2022prompt} + NPI \cite{miyake2023negativeprompt} & 73 \% & 73 \% \\
PnP \cite{Tumanyan_2023_CVPR} & 50 \% & 58 \% \\
EDICT \cite{10204740} & 80 \% & 67 \% \\
ProxMasaCtrl \cite{han2023improving} & 67 \% & 81 \% \\
MasaCtrl \cite{cao_2023_masactrl} & 29 \% & 71 \% \\
        \bottomrule
    \end{tabular}
    \label{tab:user_study_a2a}
\end{table}

\begin{table}[h!]
    \centering
    \caption{User studies that show the percentage of users that preferred our method. The current table shows the comparison on the subset of results that belong to \textbf{stylisation} editing type.}
    \small
    \begin{tabular}{lccc}
    \toprule
        \textbf{\quad \ \ \ Method}         & Preservation  & Editing \\
\midrule
P2P \cite{hertz2022prompt} + NTI \cite{mokady2022null} & 33 \% & 52 \% \\
P2P \cite{hertz2022prompt} + NPI \cite{miyake2023negativeprompt} Prox \cite{han2023improving} & 64 \% & 36 \% \\
P2P \cite{hertz2022prompt} + NPI \cite{miyake2023negativeprompt} & 46 \% & 62 \% \\
PnP \cite{Tumanyan_2023_CVPR} & 69 \% & 62 \% \\
EDICT \cite{10204740} & 50 \% & 33 \% \\
ProxMasaCtrl \cite{han2023improving} & 50 \% & 83 \% \\
MasaCtrl \cite{cao_2023_masactrl} & 76 \% & 92 \% \\
        \bottomrule
    \end{tabular}
    \label{tab:user_study_style}
\end{table}

\begin{table}[h!]
    \centering
    \caption{User studies that show the percentage of users that preferred our method. The current table shows the comparison on the subset of results that belong to \textbf{person-in-the-wild} editing type.}
    \small
    \begin{tabular}{lccc}
    \toprule
        \textbf{\quad \ \ \ Method}         & Preservation  & Editing \\
\midrule
P2P \cite{hertz2022prompt} + NTI \cite{mokady2022null} & 50 \% & 58 \% \\
P2P \cite{hertz2022prompt} + NPI \cite{miyake2023negativeprompt} Prox \cite{han2023improving} & 50 \% & 50 \% \\
P2P \cite{hertz2022prompt} + NPI \cite{miyake2023negativeprompt} & 52 \% & 33 \% \\
PnP \cite{Tumanyan_2023_CVPR} & 61 \% & 33 \% \\
EDICT \cite{10204740} & 42 \% & 58 \% \\
ProxMasaCtrl \cite{han2023improving} & 73 \% & 82 \% \\
MasaCtrl \cite{cao_2023_masactrl} & 100 \% & 100 \% \\
        \bottomrule
    \end{tabular}
    \label{tab:user_study_person}
\end{table}

\end{document}